\documentclass[lettersize,journal]{IEEEtran}
\usepackage{amsmath,amsfonts}
\usepackage{algorithmic}
\usepackage{algorithm}
\usepackage{array}
\usepackage{textcomp}
\usepackage{url}
\usepackage{verbatim}
\usepackage[pdftex]{graphicx}
\usepackage{cite}
\usepackage{booktabs}
\usepackage{cellspace}
\usepackage[shortlabels]{enumitem}
\usepackage{microtype}
\usepackage{graphicx}
\usepackage{booktabs} % for professional tables
\usepackage{hyperref}

\usepackage{amsmath}
\usepackage{amssymb}
\usepackage{mathtools}
\usepackage{amsthm}
\usepackage{enumitem}

\usepackage[capitalize,noabbrev]{cleveref}
\usepackage{hyperref}
\usepackage{xr-hyper}

\makeatletter
\newcommand*{\addFileDependency}[1]{
  \typeout{(#1)}
  \@addtofilelist{#1}
  \IfFileExists{#1}{}{\typeout{No file #1.}}
}
\makeatother

\theoremstyle{plain}
\newtheorem{theorem}{Theorem}[section]

\newtheorem{lemma}[theorem]{Lemma}

\theoremstyle{definition}

\theoremstyle{remark}
\newtheorem{remark}[theorem]{Remark}

\usepackage{hyperref}

\usepackage{url}
\usepackage{graphicx,wrapfig,lipsum} 
\usepackage{times}
\usepackage{soul}
\usepackage{url}
\usepackage[utf8]{inputenc}
\usepackage[small]{caption}
\usepackage{graphicx}
\usepackage{amsthm}
\usepackage{booktabs}
\usepackage{mwe} % for blindtext and example-image-a in example
\usepackage{wrapfig}

\usepackage{setspace}
\usepackage{amsmath,amssymb,amsfonts}
\usepackage{amsmath,amsfonts,bm}

\def\eqref#1{(\ref{#1})}
\def\1{\bm{1}}
\def\gL{{\mathcal{L}}}

\def\argmin{\text{arg\,min}}

\crefformat{equation}{(#2#1#3)}
\crefrangeformat{equation}{(#3#1#4) to~(#5#2#6)}
\crefmultiformat{equation}{(#2#1#3)}%
{ and~(#2#1#3)}{, (#2#1#3)}{ and~(#2#1#3)}
\crefname{assumption}{assumption}{assumptions}
\crefname{Assumption}{Assumption}{Assumptions}
\usepackage{bbm}
\usepackage[mathscr]{euscript}
\usepackage{graphicx}
\usepackage{textcomp}
\usepackage{xcolor}

\usepackage{xspace}
\usepackage{caption}
\usepackage{subcaption}
\usepackage{bm}
\usepackage{array}
\usepackage{multirow}

\makeatletter
\newcommand{\multilines}[1]{%
	\begin{tabularx}{\dimexpr\linewidth-\ALG@thistlm}[t]{@{}X@{}}
		#1
	\end{tabularx}
}
\makeatother

\usepackage[textsize=tiny]{todonotes}

\Crefname{figure}{Fig.}{Figs.}
\Crefname{table}{Table}{Tables}
\Crefname{section}{Sec.}{Secs.}

\DeclarePairedDelimiterX{\norm}[1]{\lVert}{\rVert}{#1}
\DeclarePairedDelimiterX{\abs}[1]{\lvert}{\rvert}{#1}
\newcommand{\innProd}[1]{\left\langle#1\right\rangle}
\newcommand{\set}[1]{\left\lbrace#1\right\rbrace}

\DeclareSymbolFont{extraup}{U}{zavm}{m}{n}
\DeclareMathSymbol{\varheart}{\mathalpha}{extraup}{86}
\newcommand{\SumNoLim}[2]{\ensuremath{\sum\nolimits_{#1}^{#2}}}
\newcommand{\SumLim}[2]{\ensuremath{\sum\limits_{#1}^{#2}}}

\newcommand{\defeq}{\vcentcolon=}
\newcommand{\BigP}[1]{\ensuremath{\Bigl(#1\Bigr)}}

\newcommand{\BigC}[1]{\ensuremath{\Bigl\{#1\Bigr\}}}

\DeclareUnicodeCharacter{FB01}{fi}
\long\def\comment#1{}

\begin{document}

\title{Federated PCA on Grassmann Manifold for IoT Anomaly Detection}

\author{Tung-Anh Nguyen,
        Long Tan Le,
        Tuan Dung Nguyen, \\
        Wei Bao,~\IEEEmembership{Member,~IEEE,} 
        Suranga Seneviratne,~\IEEEmembership{Senior Member,~IEEE,} \\
        Choong Seon Hong,~\IEEEmembership{Fellow,~IEEE,}
        Nguyen H. Tran,~\IEEEmembership{Senior Member,~IEEE.}
        % <-this % stops a space
\thanks{T.-A. Nguyen, L. T. Le, N. H. Tran, W. Bao, and S. Seneviratne are with the School of Computer Science, The University of Sydney, Sydney, NSW 2006, Australia (email: tung6100@uni.sydney.edu.au, \{long.le, nguyen.tran, wei.bao,
suranga.seneviratne\}@sydney.edu.au). T. D. Nguyen is currently with the Department of Computer and Information Science, University of Pennsylvania, USA (email: joshtn@seas.upenn.edu). C. S. Hong is with the Department of Computer Science and Engineering, School of Computing, Kyung Hee University, Yongin-si 17104, Republic of Korea (email: cshong@khu.ac.kr). \comment{Part of this work was presented at IEEE INFOCOM 2023~\cite{fedpca_infocom}. (Tung-Anh Nguyen and Long Tan Le contributed equally to this work.) (Corresponding authors: Tung-Anh Nguyen and Long Tan Le.)}}% <-this % stops a space
%\thanks{Manuscript received April 19, 2021; revised August 16, 2021.}
}

% \and
% \IEEEauthorblockN{Long Tan Le, Wei Bao, Nguyen H. Tran}
% \IEEEauthorblockA{\textit{School of Computer Science} \\
% \textit{The University of Sydney}\\
% Sydney, Australia \\
% \{long.le, wei.bao, nguyen.tran\}@sydney.edu.au}
% The paper headers
\markboth{Journal of \LaTeX\ Class Files,~Vol.~14, No.~8, August~2021}%
{Shell \MakeLowercase{\textit{et al.}}: A Sample Article Using IEEEtran.cls for IEEE Journals}

%\IEEEpubid{0000--0000/00\$00.00~\copyright~2021 IEEE}
% Remember, if you use this you must call \IEEEpubidadjcol in the second
% column for its text to clear the IEEEpubid mark.

\maketitle

\begin{abstract}
With the proliferation of the Internet of Things (IoT) and the rising interconnectedness of devices, network security faces significant challenges, especially from anomalous activities. While traditional machine learning-based intrusion detection systems (ML-IDS) effectively employ supervised learning methods, they possess limitations such as the requirement for labeled data and challenges with high dimensionality. Recent unsupervised ML-IDS approaches such as AutoEncoders and Generative Adversarial Networks (GAN) offer alternative solutions but pose challenges in deployment onto resource-constrained IoT devices and in interpretability. To address these concerns, this paper proposes a novel federated unsupervised anomaly detection framework -- FedPCA -- that leverages Principal Component Analysis (PCA) and the Alternating Directions Method Multipliers (ADMM) to learn common representations of distributed non-i.i.d. datasets. Building on the FedPCA framework, we propose two algorithms, \textsc{FedPE} in Euclidean space and \textsc{FedPG} on Grassmann manifolds. Our approach enables real-time threat detection and mitigation at the device level, enhancing network resilience while ensuring privacy. Moreover, the proposed algorithms are accompanied by theoretical convergence rates even under a sub-sampling scheme, a novel result. Experimental results on the UNSW-NB15 and TON-IoT datasets show that our proposed methods offer performance in anomaly detection comparable to non-linear baselines, while providing significant improvements in communication and memory efficiency, underscoring their potential for securing IoT networks.
\end{abstract}


\section{Introduction}
\IEEEPARstart{T}{he} ubiquity of Internet of Things (IoT) technologies has led to a transformative shift where intelligent devices exchange information and enact decisions via the internet, requiring minimal human intervention. This has significantly enhanced the quality of life and found applications across numerous sectors such as smart city infrastructures~\cite{Zanella2014}, energy systems~\cite{Yang2016}, or climate systems~\cite{Afroz2018}. However, as the IoT ecosystem expands and becomes increasingly complex, ensuring robust network security emerges as a paramount task, especially considering the inherent security vulnerabilities of many IoT devices due to their limited computational resources~\cite{Cook2020}.

In this intricate landscape, machine learning-based intrusion detection systems (ML-IDS) form a crucial line of defense. These systems often utilize supervised learning methods that range from traditional techniques to more sophisticated deep learning approaches~\cite{survey_mlids}. While effective in various scenarios, they come with several limitations. For instance, they necessitate access to a considerable volume of \emph{labeled data}, which can be time-consuming and costly to generate. Additionally, they often struggle with \emph{high-dimensional data} and are limited in their ability to detect novel anomalies not represented in the training data.  In contrast, recent unsupervised ML-IDS methods, such as AutoEncoders~\cite{ae1} and GAN-based models~\cite{bigan, bigan1}, provide anomaly detection by learning representation of normal behavior without relying on labeled data. However, these techniques often involve training deep neural network models, which can be computationally demanding and lack clear interpretability~\cite{adreview2} -- factors that pose significant challenges for resource-limited IoT devices. 

Traditionally, most ML-IDS lean on centralized deployment which is becoming less feasible in IoT networks due to privacy concerns and resource limitations. %\red{Moreover, the distributed nature of IoT environments amplifies security threats as each device serve as a potential point of vulnerability}. 
Hence, there's a heightened demand for distributed security frameworks where IoT devices can have self-monitoring capabilities that allow them to identify potential threats in their network traffic while contributing their insights to the broader system. Such a distributed approach would considerably enhance the resilience of IoT networks, enabling real-time anomaly detection and mitigation in heterogeneous environments. In light of this, Federated Learning (FL) has stepped in as a promising paradigm that allows IoT devices to collaboratively learn a shared model while retaining their data locally, thereby harmonizing privacy preservation and efficient resource utilization. \emph{Despite its potential, the exploration of federated unsupervised IDS under the heterogeneity of IoT traffic remains relatively nascent}, indicating the need for further research in this area.

In response to these challenges, we introduce a novel federated unsupervised anomaly detection framework that leverages the representation power of PCA~\cite{pca} and the decomposability of ADMM~\cite{boyd2011distributed}, namely FedPCA~\cite{fedpca_infocom}. Specifically, we aim to harness PCA's capabilities to learn a common representation of distributed non-i.i.d datasets via consensus optimization that captures the hidden states within the data. These hidden states typically reflect complex correlations and underlying patterns within the data, which are instrumental in effective anomaly detection. In order to solve FedPCA, we design two distinct algorithms including FedPCA in Euclidean space (\textsc{FedPE}) and FedPCA on Grassmann manifolds (\textsc{FedPG}). Each algorithm possesses unique properties offering potential advantages for FL. While \textsc{FedPE} operates within a Euclidean space affording both simplicity and interpretability, \textsc{FedPG} projects the data onto Grassmann manifolds providing an efficient way to handle the high-dimensional, complex, and potentially exhibit non-linear structures of network data. We then investigate their convergence characteristics and provide convergence guarantees under standard assumptions. Our proposed FL framework offers notable advantages: it preserves privacy while providing efficient anomaly detection in resource-constrained environments. 

Within this work, each IoT device functions as an anomaly detection system, playing an active role in fortifying the system's security.  The devices continuously monitor their communication patterns, employing anomaly detection models to identify abnormal behaviors that could signify a potential threat. By deploying ML-IDS at the IoT device level, we establish a distributed network of vigilant sentinels, each capable of real-time threat identification. It enables IoT devices to partake in the learning process without sharing sensitive local data—a characteristic of paramount importance in an IoT setting where user data privacy is highly valued. This setup substantially enhances the resilience and security of the entire IoT network, and requires a shorter detection time compared to its unsupervised counterparts. 
%While PCA is a well-known technique excelling in dimensionality reduction, it is capable of anomaly detection capturing hidden states within data and requires a shorter detection time compared to its supervised counterparts.
The main contributions of this work can be summarized as follows:
\begin{itemize}%[topsep=0pt]
	\setlength{\parskip}{5pt}
    \item We propose an unsupervised federated PCA framework for efficient host-based IoT anomaly detection, which is formulated by a consensus optimization problem for privacy-preserving and communication-efficient.
    \item We introduce a new algorithm designed for \textsc{FedPE} and \textsc{FedPG} based on ADMM-based procedures with client sub-sampling schemes aiming to enhance robustness and mitigate communication bottlenecks. 
    \item We provide the theoretical convergence analysis for the proposed framework, showing that it converges to a stationary point where the consensus constraint is satisfied.
    \item Through experiments on the UNSW-NB15~\cite{unsw} and TON-IoT network~\cite{ton} datasets, we demonstrate that \textsc{FedPE} and \textsc{FedPG} not only show competitive performance against non-linear methods in anomaly detection but also boast significant improvements in communication and memory efficiency. The extensive evaluation showcases the robustness of our proposed framework to non-i.i.d. data distribution, marking it well-suited for IoT networks.
\end{itemize}

The rest of the paper is organized as follows. Section~\ref{sec:background} outlines the relevant background and previous works that closely relate to our topics of interest. The problem formulation and the proposed framework are introduced in Section~\ref{sec:algorithm} with a detailed explanation of algorithm designs. In Section~\ref{sec:convergence_analysis}, we present theoretical convergence analyses for the proposed algorithms. Numerical results and conclusion are conducted in Section~\ref{sec:experiment} and Section~\ref{sec:conclusion}, respectively.
\section{Related Works}
\label{sec:background}
\subsection{Federated Learning} 
Federated Learning (FL) represents a cutting-edge distributed learning approach that allows for the execution and subsequent updating of machine learning models across thousands of participating devices, while simultaneously safeguarding against data leaks~\cite{Konecny2015}. In FL, models are deployed and trained locally on individual clients' devices using local data. These locally trained models are then transmitted back to the central server for aggregation, such as weight averaging~\cite{fedavg}. Thanks to the attributes of efficient communication and privacy protection, FL proves highly advantageous for IoT networks and applications. Various versions of FL have been developed after the birth of the standard FedAvg~\cite{fedavg}, significantly augmenting the intelligence of IoT systems across a multitude of applications~\cite{fl_iot}.

The primary challenges of FL can be broadly classified into two categories: \textit{communication bottlenecks}, resulting from the need to exchange large model's parameters between parties, and \textit{data heterogeneity}, arising from the diversity of the data held by FL clients. This motivated some prior works to design more communication-efficient strategies such as local updates \cite{konecny_federated_2017,suresh_distributed_2017, pmlr-v108-reisizadeh20a}, model compression \cite{com1}, and adaptive communication \cite{comadaptive}. Meanwhile, data heterogeneity concerned with clients' non-i.i.d. data has also been well-studied \cite{zhao_federated_2018, li_convergence_2020}.
\vspace{-5pt}
\subsection{Optimization on Grassmann Manifolds}
Grassmann manifold optimization has recently gained traction as an advanced mathematical tool in the fields of machine learning. This approach leverages the geometric properties of Grassmann manifolds, a space representing linear subspaces, to facilitate the analysis and understanding of high-dimensional data, thereby enabling efficient optimization and enhanced modeling in various machine learning and data analysis tasks~\cite{zhang2018grassmannian}. In centralized learning, the optimization over Grassmann manifolds has been employed to enhance manifold learning and data classification, with applications extending to computer vision and speech recognition~\cite{manifold1, manifold2, manifold3}. In decentralized learning contexts, particularly in FL, Grassmann learning has not fully explored yet, making it a promising avenue for continued research and development.
\vspace{-10pt}
\subsection{IoT Anomaly Detection} 
Owing to their inherent vulnerabilities, IoT systems require robust measures against malicious activities~\cite{Neshenko2019}. This need has propelled anomaly-based Intrusion Detection Systems (IDS) into the limelight, bolstered by the radical advancements in machine learning. Early research on anomaly-based IDS within IoT networks has been heavily centered around methodologies where data is consolidated onto a central server~\cite{Meidan2018, Anthi2019, Pajouh2019, Zolanvari2019, Wang2021, Qiu2021}, raising concerns related to privacy and cyber intrusions.

The recent trend has veered towards the use of FL frameworks for network anomaly detection. For example, Nguyen et al.\cite{Nguyen2019} developed a federated intrusion detection system using Recurrent Neural Network models that were trained on local gateways and then amalgamated into a single model. Similarly, Wang et al.\cite{Wang2022} demonstrated how a feed-forward artificial neural network can augment detection performance. Mothukuri et al.\cite{Mothukuri2022} employed federated training on Gated Recurrent Units (GRUs) models, preserving data on local IoT devices by only sharing the learned weights with the central server. However, accurately detecting unknown anomalous activities has proven challenging, given the inherent imbalance between normal and abnormal traffic in IoT networks\cite{Cook2020}.
\vspace{-10pt}
\subsection{Principal Component Analysis} 
Over recent years, PCA~\cite{Jolliffe2014} has become a popular unsupervised learning model in centralized machine learning. From an IoT security standpoint, PCA-based anomaly detection has emerged as an efficient technique, with PCA deployed to bifurcate a dataset into benign and malicious subspaces~\cite{Lakhina2004}. For instance, Brauckhoff et al. ~\cite{Brauckhoff2009} suggested the use of the Karhunen-Loeve expansion in light of the data's temporal correlation, albeit limited to data from a single router.
Kernel PCA has been put forward to examine features considering both variance and mean~\cite{Chen2021}. However, these methods, while promising, are constrained by their reliance on data from a single device, rendering them inadequate for IDS in settings where resources are limited and privacy is paramount.

Distributed PCA has gained traction in the machine learning community~\cite{Wu2018}. For example, Ge et al.\cite{pmlr-v84-ge18a} proposed a minimax optimization framework for generating a distributed sparse PCA estimator while maintaining privacy. Grammenos et al.\cite{Grammenos2020} suggested an asynchronous FL method to implement PCA in a memory-limited environment, introducing an algorithm that calculates local model updates using a streaming process and adaptively estimates its principal components.

% Recently, the idea of distributed PCA has become prevalent in the field of Machine Learning.
 
%~\cite{pmlr-v84-ge18a}
%~\cite{Grammenos2020}

%  Therefore, an unsupervised approach that profiles normal data and detects anomaly when they fall outside the profile would be used for this work.
\vspace{-10pt}
\subsection{ADMM-based Learning} 

The Alternating Direction Method of Multipliers (ADMM) has made significant advancements in both theory and application over recent decades~\cite{boyd2011distributed}. ADMM acts as a powerful framework that combines the strengths of dual decomposition and the method of multipliers to solve optimization problems. The core idea behind this framework is to break down a problem into simpler subproblems that can be solved in parallel, and then combine the solutions of these subproblems to arrive at a global solution. The mathematical and convergence characteristics of ADMM have been well-studied in the literature~\cite{giesen2016distributed, Hong2015,Chang2016, admm_nonconvex}. 

Owing to its decompositional nature, ADMM has found significant applications in large-scale and distributed optimization tasks, especially in machine learning~\cite{admm2, admm3}.  In FL, several ADMM-based decentralized methods have been developed to enhance communication efficiency~\cite{admm1, admm4}. However, \emph{employing this framework for distributed PCA is still uncharted}. 

In this study, we employ an alternative strategy to develop an effective unsupervised method for network anomaly detection in IoT systems. We propose an unsupervised federated PCA framework grounded in ADMM-based learning, addressing privacy concerns while efficiently detecting anomalies in both recognized and unprecedented attacks.

\section{Problem Formulation}
\label{sec:algorithm}
\begin{figure}[t]
	\centering    
	\includegraphics[width=1.0\columnwidth]{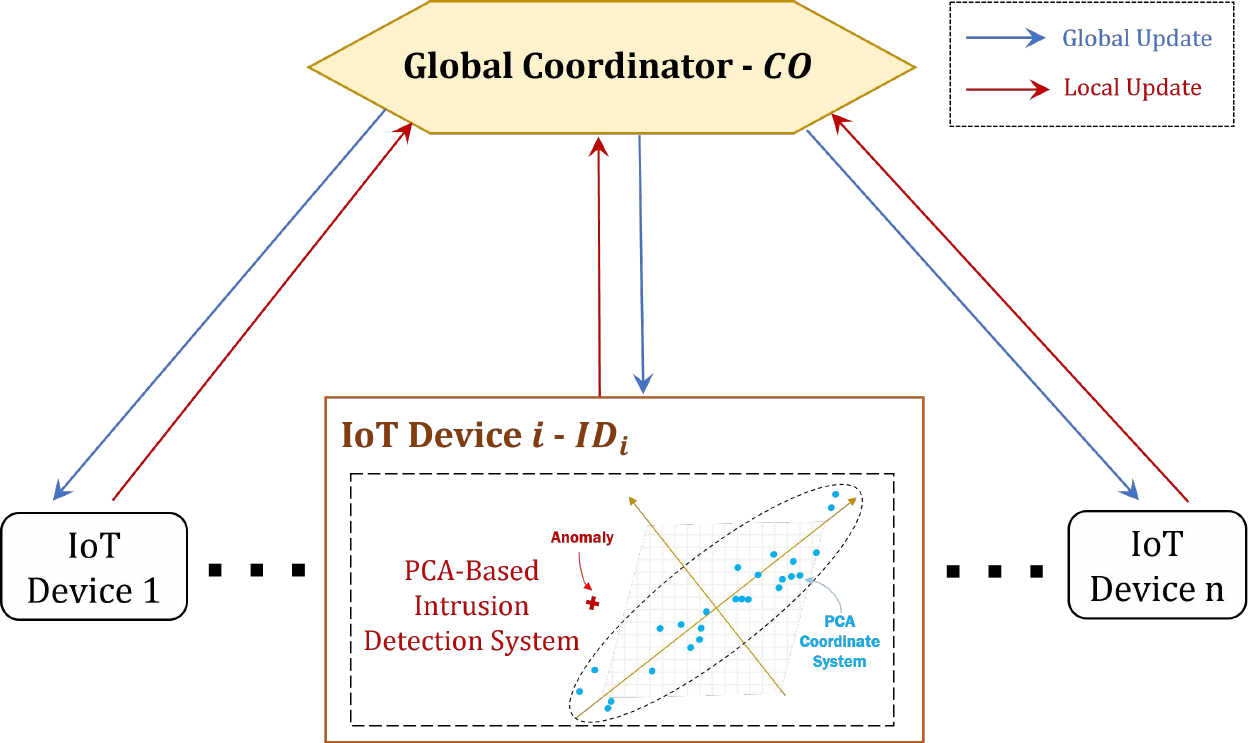}
	\caption{Overview of a host-based IoT anomaly detection system utilizing Federated PCA.}
	\label{fig:system}
 \vspace{-10pt}
\end{figure}

\subsection{Federated Principal Component Analysis}

We consider an IoT network comprised of a central global coordinator $CO$ and a set of $N$ IoT devices, each denoted by a unique identifier $ID_i$ holding a local data set $X_i$, where $i \in {1,2,...,N}$. This network interacts within a client-server topology, as depicted in Fig. \ref{fig:system}. Our framework employs a host-based IDS strategy, where each IoT device operates as a sentinel within the network, examining its network traffic and identifying potential security threats. This distributed approach significantly increases the network's resilience by facilitating real-time threat detection and mitigation. Furthermore, since the learning is local to each device, it allows for a reduction in communication overhead and strengthens privacy by keeping the data on the device.

\vspace{5pt}
\subsubsection*{\textbf{Principal Component Analysis Revisiting}} We first revisit the basics of PCA to formally define our method. PCA projects a given dataset $X$ onto principal components ordered by the amount of variance they capture in the data~\cite{Jolliffe2002}. Specifically, it seeks an optimal rank-$k$ ($k < d$) matrix $\tilde{X}$ that minimizes the reconstruction error $\norm{X - \tilde{X}}_F$, where $||\cdot||_F$ denotes the Frobenius norm~\cite{Udell2016}. Typically, $\tilde{X}$ is derived based on the singular value decomposition (SVD) of $X$, i.e.,
% \begin{align*}
	$X = U \Sigma V^\top$,
% \end{align*}
where $U \in \mathbb{R}^{d \times d}$ is an orthonormal matrix containing the left singular vectors of $X$, $\Sigma = \text{diag}(\sigma_1, \ldots, \sigma_d)$ contains the singular values ($\sigma_1 \geq \sigma_2 \geq \ldots \geq \sigma_d$), and $V \in \mathbb{R}^{D \times D}$ contains the right singular vectors of $X$. Let $U^{(k)} \in \mathbb{R}^{d \times k}$   be the first $k$ columns of $U$ and $\Sigma_k = \text{diag}(\sigma_1, \ldots, \sigma_k)$. We can rewrite the solution for $\tilde{X}$ as
\begin{equation}
\label{E:pca_centralized}
	\tilde{X} = U^{(k)} U^{(k)\top} X,
\end{equation} 
which can be interpreted as the projection of data matrix $X$ onto the column space of $U^{(k)}$. 
The PCA problem can thus be formulated as an optimization problem as follows:
\begin{equation} \label{Prob:pca_centralized}
	\begin{aligned}
		\min_{U \in \mathbb{R}^{d \times k}} \quad & \norm{(I - U U^\top)X}_F^2 \\
		\mathrm{s.t.} \quad & U^\top U = I.
	\end{aligned}
\end{equation}
The essence of this problem is to locate an orthogonal matrix $U$ that minimizes the reconstruction error of $X$ when projected onto the column space of $U$, resulting in a matrix $Z = U^\top X \in \mathbb{R}^{k \times D}$ provides a lower-dimensional representation of $X$. 

In the context of developing ML-IDS for IoT networks, solving this PCA problem typically requires IoT devices to transmit their local data to a centralized server. While this approach offers a streamlined method for anomaly detection, it is hampered by two main obstacles.  
First,  data transferring within the resource-limited and bandwidth-constrained landscape of IoT can lead to significant \textit{communication bottlenecks}, which in turn might escalate operational costs, amplify latency, and induce network congestion. 
Second, the act of centralizing such data amplifies \textit{privacy concerns} as it risks exposing sensitive information from personal environments, posing serious threats to user confidentiality. Hence, an alternative would be to seek distributed methods that address these issues while still effectively detecting anomalies. %The content for this answer is also reflected by section $V.B.1$ in the manuscript. 
% \vspace{-44.4pt}
\subsubsection*{\textbf{Federated PCA via ADMM Consensus Optimization}} To address these concerns, we propose a federated approach for solving PCA on distributed datasets, namely FedPCA, which aims to collectively learn a \emph{common representation} from all local datasets, $X_1, X_2, ..., X_N$, without the $CO$ directly accessing these datasets. We first introduce a set of local variables $U_1, \ldots, U_N$, where $U_i$ is the local PCA matrix learned by client $i$. Then, we formulate a consensus optimization problem as follows:
\begin{equation} \label{Prob:pca_decentralized_admm}
	\begin{aligned}
		\min_{U_1, \ldots, U_N} \quad & { \SumLim{i=1}{N} \norm{(I - U_i U_i^\top)X_i}_F^2} \\
		\mathrm{s.t.} \quad & U_i = Z, \; \forall i=1,\ldots,N. \\
		 \quad & U_i^\top U_i = I, \; \forall i=1,\ldots,N, 
	\end{aligned}
\end{equation}
where the first linear constraint enforces the ``consensus'' to a \emph{common low-rank matrix} $Z$ \cite{Boyd2011} for all local $U_i$. The second constraint ensures that the matrix $U_i$ is orthonormal for every client keeping the distributed problem consistent with the classical centralized PCA. However, addressing the non-linearity introduced by this orthonormality constraint is a challenging task due to the non-convex nature of the problem. Here, we employ a surrogate technique, as introduced in \cite{giesen2016distributed}, to handle the above non-linear constraint as follows. 
\begin{equation}
\label{eqn:Hi}
	h_i (U_i) = \max{\set{0, U_i^\top U_i - I}}^2,
\end{equation}
where the operators $\max$ and squared are component-wise. By reformulating the problem to the $h_i (U_i)$ function, we gently steer the optimization towards matrices $U_i$ that approach orthonormality, making the landscape more tractable for iterative frameworks. We then have the problem \eqref{Prob:pca_decentralized_admm} equivalent to the following:
\begin{equation} \label{Prob:pca_decentralized_admm_nonlinear}
	\begin{aligned}
		\min_{U_1, \ldots, U_N} \quad & { \SumLim{i=1}{N} \norm{(I - U_i U_i^\top)X_i}_F^2} \\
		\mathrm{s.t.} \quad & U_i = Z, \; \forall i=1,\ldots,N \\
		 \quad & h_i(U_i) \leq 0, \; \forall i=1,\ldots,N.
	\end{aligned}
\end{equation}
Given the nature of problem \eqref{Prob:pca_decentralized_admm_nonlinear} as a consensus optimization with linear constraints, we opt to utilize ADMM~\cite{boyd2011distributed} for its resolution. The essence of this iterative framework is the alternating updates of primal and dual variables to achieve two main objectives: the $U_i$ across clients converge towards a common $Z$ and the objective function is iteratively minimized. Applying this to our problem, we first construct the augmented Lagrangian function associated with the problem~\eqref{Prob:pca_decentralized_admm_nonlinear} as follows.
\vspace{-10pt}
\begin{equation}
\begin{split}
	&\gL(\{U_i\}, Z, \{Y_i\}, \{T_i\}) = \SumLim{i=1}{N} f_i(U_i) + \SumLim{i=1}{N}\innProd{Y_i, U_i - Z}_F \\ & + \SumLim{i=1}{N}\innProd{T_i, h_i(U_i)}_F + \frac{\rho}{2} \SumLim{i=1}{N} \norm{U_i - Z}_F^2 + \frac{\rho}{2} \SumLim{i=1}{N} \norm{h_i(U_i)}_F^2,
\end{split}
\label{eq:augLag}
\end{equation}
where $f_i(U_i) = \norm{(I - U_i U_i^\top)X_i}_F^2$ is the objective function, $\innProd{\cdot, \cdot}_F$ is the Frobenius inner product between two matrices.

This augmented Lagrangian makes the optimization problem more amenable to iterative solution techniques while ensuring that constraints are respected. The terms with Lagrange multipliers (or dual variables) -- $\SumNoLim{i=1}{N}\innProd{Y_i, U_i - Z}_F$ and  $\SumNoLim{i=1}{N}\innProd{T_i, h_i(U_i)}_F$ -- quantify the influence of the constraints on achieving a common PCA representation across distributed datasets. They ensure that each local PCA matrix $U_i$ converges to a shared matrix $Z$ and adheres to the non-linear constraints dictated by $h_i(U_i)$. On the other hand, the penalty terms -- $(\rho/2) \SumNoLim{i=1}{N} \norm{U_i - Z}_F^2$  and $(\rho/{2} )\SumNoLim{i=1}{N} \norm{h_i(U_i)}_F^2$ -- are governed by the penalty parameter $\rho$ to penalize any deviation of the local PCA matrices $U_i$ from the global consensus matrix $Z$ or from satisfying the orthogonality conditions $h_i(U_i)$. 

Since, the Frobenius inner product for matrices is analogous to the standard dot products for vectors, we can obtain the gradients of the augmented Lagrangian \eqref{eq:augLag} w.r.t. dual variables $Y_i$ and $T_i$ as follows.
$$\nabla_{Y_i} \mathcal{L}_{\rho}(U, Z, Y, T) = U_i - Z$$  $$\nabla_{T_i} \mathcal{L}_{\rho}(U, Z, Y, T) = h_i(U_i)$$
% \begin{align*}
% 	\nabla_{Y_i} \mathcal{L}_{\rho}(U, Z, Y, T) &= U_i - Z \\
% 	\nabla_{T_i} \mathcal{L}_{\rho}(U, Z, Y, T) &= h_i(U_i).
% \end{align*}

\begin{algorithm}[t]
    %\small
	\caption{FedPCA in Euclidean Space (\textsc{FedPE})}
	\label{algo1}
	\begin{algorithmic}[1]
% 		\State \tbf{Input:} $U$
		\STATE Randomly initialize $Z^0$ and $U_i^0, \quad \forall i = 1, \dots, N$
		\FOR{$k = 0, \ldots, T - 1$}  %\COMMENT{\textit{Global rounds}}
		%\State Broadcast $Z^{k}$ to all clients
        \STATE Sample subset clients $\mathcal{S}^k$
		  \FOR {each client $i \in \mathcal{S}^k$ in parallel}
        
        %\COMMENT{\textit{Local rounds}} 
%  		\State $U^k_i = Z^k$
% 		\State Update $U_i$ 
% 		\begin{align*}
% 			U_i^{k+1}= &\argmin_{U_i} \BigC{f_i(U_i) + \innProd{Y_i^k, U_i - Z^k}_F \\ &+ \innProd{T_i^k, h_i(U_i)}_F + \frac{\rho}{2} \norm{U_i - Z^k}_F^2 + \frac{\rho}{2} \norm{h_i(U_i)}_F^2}
% 		\end{align*}
		\STATE $U_i^{k+1} = \underset{U_i}{\operatorname{\argmin}}  \BigC{f_i(U_i) + \innProd{Y_i^k, U_i - Z^k}_F + \innProd{T_i^k, h_i(U_i)}_F + \frac{\rho}{2} \norm{U_i - Z^k}_F^2 + \frac{\rho}{2} \norm{h_i(U_i)}_F^2}$
		\ENDFOR
% 		\State Aggregate all $U_i^{k+1}$ from clients
		\STATE  $CO$ updates $Z^{k+1} = \frac{1}{|\mathcal{S}^k|} \sum_{i \in \mathcal{S}^k}{} (U_i^{k+1} +  \frac{1}{\rho} Y_i^k$)  %\COMMENT{\textit{Global update}}
		\STATE $CO$ broadcasts $Z^{k+1}$ to all clients
		\FOR {each client $i \in \mathcal{S}^k$ in parallel} %\COMMENT{\textit{Local update}}
		\STATE   $Y_i^{k+1} = Y_i^k + \rho \BigP{U_i^{k+1} - Z^{k+1}} $ %\Comment{Local update}
		\STATE   $T_i^{k+1} = T_i^k + \rho \, h_i(U_i^{k+1})$ %\Comment{Local update}
		\ENDFOR
		\ENDFOR
	\end{algorithmic}
 
\end{algorithm}

To solve the proposed problem, we employ an ADMM-based procedure in which a subset of clients, denoted as $\mathcal{S}^k$, performs iterative updates using a step size $\rho$ during the $k$-th communication round as follows.

\subsubsection{Primal Update (Local Update)}
\begin{align}
	U_i^{k+1} &= \underset{U_i}{\text{argmin}} \BigC{f_i(U_i) + \innProd{Y_i^k, U_i - Z^k}_F + \innProd{T_i^k, h_i(U_i)}_F \nonumber \\ 
		& \quad \quad \quad \quad \quad  + \frac{\rho}{2} \norm{U_i - Z^k}_F^2 + \frac{\rho}{2} \norm{h_i(U_i)}_F^2}
  \label{ADMM_PCA_update_U_i}
\end{align}

\subsubsection{Consensus Update (Global Update)}
\begin{equation}
    %Z^{k+1} &= \frac{1}{|\mathcal{S}_k|} \sum_{i \in \mathcal{S}_k} U_i^{k+1}.
    Z^{k+1} = \textcolor{black}{\frac{1}{|\mathcal{S}_k|} \sum_{i \in \mathcal{S}_k}} \BigP{U_i^{k+1} + \frac{1}{\rho} Y_i^k}
    \label{ADMM_PCA_update_Z}
\end{equation}

\subsubsection{Dual Update (Local Updates)}
\begin{equation}
    Y_i^{k+1} = Y_i^k + \rho \BigP{U_i^{k+1} - Z^{k+1}} \label{ADMM_PCA_update_Y_i}
\end{equation}
\begin{equation}
    T_i^{k+1} = T_i^k + \rho \, h_i(U_i^{k+1}). \label{ADMM_PCA_update_T_i}
\end{equation}
% In case of $\mathcal{S}_k = N$, substituting $Z^{k+1}$ to $Y_i^{k+1}$ leads to $\textcolor{blue}{\frac{1}{N} \sum_{i=1}^{N}} Y_i^{k+1}=0$, which means $Z$ update can be rewritten as \cite{Boyd2011}:

When $|\mathcal{S}_k| = N$, substituting $Z^{k+1}$ to $Y_i^{k+1}$ results in $\frac{1}{N} \sum_{i=1}^{N} Y_i^{k+1}=0$. This which means $Z$ update can be rewritten as\cite{Boyd2011}:
\begin{align}
    Z^{k+1} &= \frac{1}{N} \sum_{i=1}^{N} U_i^{k+1}.
    \label{ADMM_PCA_update_Z_2}
\end{align}

These steps are repeated iteratively until convergence, at which point the local PCA matrix $U_i$ and the consensus variable $Z$ agree, resulting in a common matrix learned from the distributed data. This ADMM-based framework collaboratively drives towards a consensus PCA solution while ensuring that the unique characteristics of each dataset are preserved.

The above procedure is denoted as FedPCA in Euclidean space, abbreviated as \textsc{FedPE}.  The full procedure is detailed in Alg.~\ref{algo1}. Specially, in each round, a distinct subset of clients $\mathcal{S}^k$ is selected to perform the update (Alg.~\ref{algo1}, line 3). Each selected client is then involved in solving the primal problem \eqref{ADMM_PCA_update_U_i} on Euclidean space locally using gradient-based methods (Alg.~\ref{algo1}, line 5). Once finishing training, selected clients send their local PCA matrix to the $CO$ for updating the common PCA matrix $Z$ (Alg.~\ref{algo1}, line 7). Finally, each selected client i update its dual variable $Y_i$ and $T_i$ based on the latest $U_i$ and $Z$ (Alg.~\ref{algo1}, lines 10-11). One main difference between \textsc{FedPE} in this work and the one presented in \cite{fedpca_infocom} is that we employ a client subsampling strategy during the federated training process. This offers significant benefits, including reduced communication overhead, improved scalability for large-scale IoT networks, and enhanced resilience to stragglers or non-responsive devices.
\begin{figure}[t]
% 	\vspace{-10pt}
	\centering    \includegraphics[width=0.6\columnwidth]{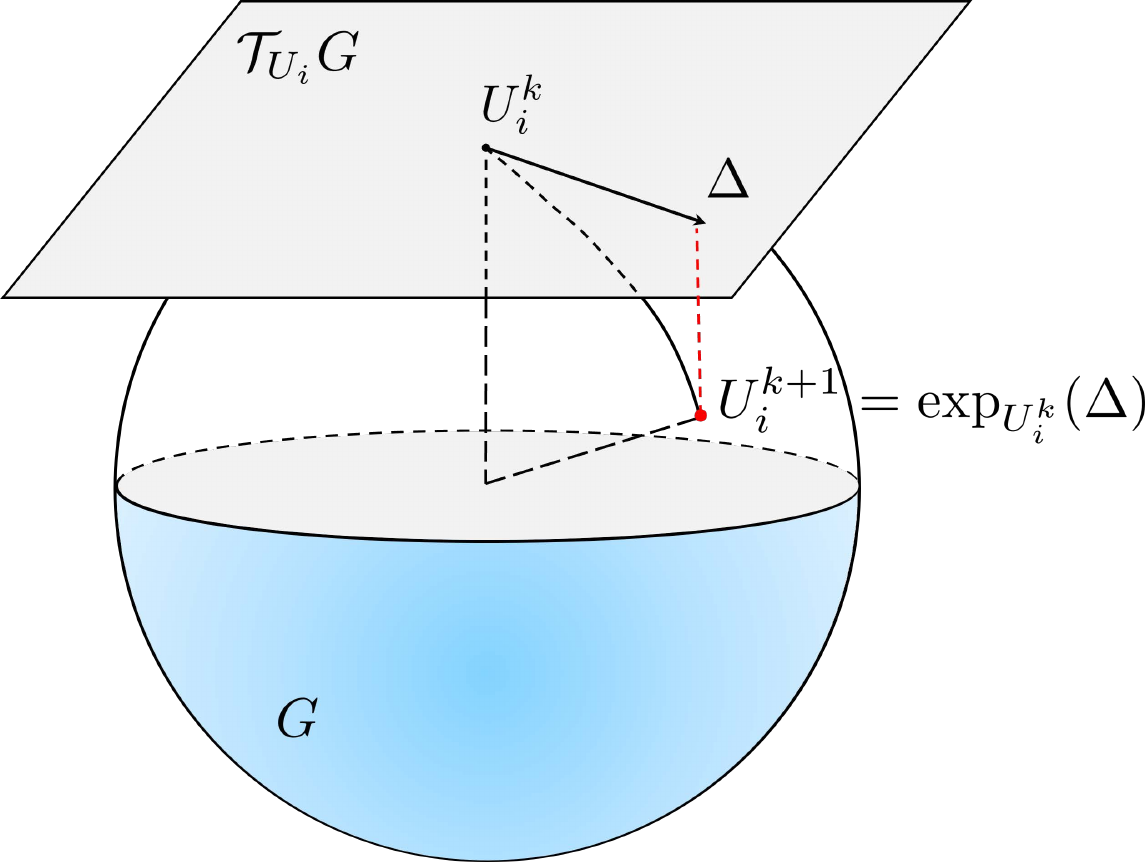}
% 		\vspace{-10pt}
	\caption{Illustration of movement on a Grassmann manifold, represented by a sphere. Given a point $U_i^k$ on Grassmann manifold $G$ and a vector $\Delta$ on the tangent space denoted by $\mathcal{T}_{U_i}G$ at $U_i^k$, a point $U_i^{k+1}$ is identified by exponential mapping $\exp_{U_i^k}(\Delta)$.  In our context, the manifold $G$ contains all matrix $U_i$ satisfy the condition $U_i^{\top}U_i = I$. }
	\label{fig:gm}
 %\vspace{-10pt}
\end{figure}

\vspace{-19.2pt}
\subsection{FedPCA using ADMM on Grassmann Manifold}~\label{FedPG}
Although \textsc{FedPE} effectively addresses the proposed problem, we further extend its capacity by formulating FedPCA on Grassmann manifolds, denoted as \textsc{FedPG}, to accelerate convergence. Notably, the orthonormality constraint utilized in~(\ref{Prob:pca_decentralized_admm}) signifies that the parameters of (\ref{ADMM_PCA_update_U_i}) can be projected onto a Grassmann manifolds~\cite{zhang2018grassmannian}.

\subsubsection*{\textbf{Orthonormality meets the Grassmannian}} The Grassmann manifold, denoted as $G(n,d)$, represents a geometric space that encapsulates sets of $d$-dimensional subspaces within an overaching $n$-dimensional real or complex Euclidean space, subject to $n \geq d > 0$~\cite{zhang2018grassmannian}. The distinct of this manifold lies in its representation capacity: Any point on $G(n,d)$ can be identified by an $n\times d$ orthogonal matrix $U$ whose column spans the corresponding subspace to ensure orthonormality $U^\top U= I_d$. Hence, $G(n,d)$ is defined as follows.
\begin{equation}\label{eq:grass-constraint}
 	G(n, d) = \{ \text{span}(U): U \in \mathbb{R}^{n\times d}, U^\top U= I_d \}
\end{equation}

Fig.~\ref{fig:gm} shows an example of the Grassmann manifold as the surface of a sphere. Here, the tangent space $\mathcal{T}_U G$ at $U$ captures the set of all directions in which the subspace can "move" but still stay very close to the manifold. By capturing the space of all orthonormal subspaces, the Grassmann manifold offers an elegant framework for addressing optimization problems with orthonormality constraints (e.g., problem \eqref{Prob:pca_decentralized_admm}) since these constraints are automatically maintained when operations are performed within this manifold.
%To achieve this,

\comment{
Consider a function $F: {G(n,k) \rightarrow \mathbb{R}}$ and its Euclidean gradient $F_A$ at point $A$, 
the projected gradient of the function $F$ on the Grassmann manifold is computed by projecting the Euclidean gradient $F_A$ onto the tangent space of the Grassmann manifold using the orthogonal projection $F_A \rightarrow \nabla_A{F}$:
\begin{equation}
	\nabla_AF\ = (I_k - AA^{\top})F_A.
\end{equation}
}
%To solve a gradient-based learning problems in Grassmann manifold, the notion of tangent space is required. Given a point $A\in G(n,k)$, tangent space is defined as a set of vectors $\{\nabla\}$ such that $A^{\top}\nabla = 0$.  Intuitively, tangent space is a real vector space that contains the possible directions in which one can tangentially pass through $A$.

\subsubsection*{\textbf{FedPCA via Grassmann Manifold Optimization}}
By framing problem \eqref{Prob:pca_decentralized_admm} within the Grassmann manifold, the augmented Lagrangian \eqref{eq:augLag} evolves as follows.
\begin{equation}
\begin{split}
    \label{eqn: lagrange}
	& \gL(\{U_i\}, Z, \{Y_i\})  \\ & = \SumLim{i=1}{N} f_i(U_i) + \SumLim{i=1}{N} Y_i^\top (U_i - Z) + \frac{\rho}{2} \SumLim{i=1}{N} \norm{U_i - Z}^2_F.
\end{split}
\end{equation}
Here, the terms associated with $h_i(U_i)$ are eliminated because the inherent orthonormality constraints are satisfied. 

Denoting the local update for each client $i$ at the $k$-th iteration by following:
\begin{equation} \label{E:Fopt_2}
	%U_i^{k+1} &= \argmin_{U_i} \BigC{F_i^k(U_i)} \quad \mathrm{s.t.}
	F_i^k(U_i) := f_i(U_i) + Y_i^{k\top}(U_i - Z^k) + \frac{\rho}{2} \norm{U_i - Z^k}_F^2
% 	Z^{k+1} &= \frac{1}{N} \SumLim{i=1}{N} \bigP{U_i^{k+1} + \frac{1}{\rho}Y_i^k}. \\ 
% 	Y_i^{k+1} &= Y_i^k + \rho \bigP{U_i^{k+1} - Z^{k+1}}.
\end{equation}

We then recast the update rules for $U_i$ and $Z$ on the Grassmann manifold as follows.
\begin{equation}
    \label{E:Fopt}
	U_i^{k+1} = \argmin_{U_i} \BigC{F_i^k(U_i)} \quad %\mathrm{s.t.}
	%F_i^k(U_i) &:= f_i(U_i) + Y_i^{k\top}(U_i - Z^k) + \frac{\rho}{2} \norm{U_i - Z^k}^2.
% 	Z^{k+1} &= \frac{1}{N} \SumLim{i=1}{N} \bigP{U_i^{k+1} + \frac{1}{\rho}Y_i^k}. \\ 
% 	Y_i^{k+1} &= Y_i^k + \rho \bigP{U_i^{k+1} - Z^{k+1}}.
\end{equation}

\begin{algorithm}[t]
    % \algsetup{linenosize=\tiny}
    %\small
	\caption{FedPCA on Grassmann Manifold (\textsc{FedPG})}
    \label{algo2}
	\begin{algorithmic}[1]
% 		\State \tbf{Input:} $U$
		\STATE Randomly initialize $Z^0$ and $U_i^0, \quad \forall i = 1, \dots, N$
		\FOR{$k = 1, \ldots, T$} %\Comment{\textit{Global  rounds}}
	%	\State Broadcast $Z^{k}$ to all clients
        \STATE Sample subset clients $\mathcal{S}^k$
		  \FOR {each client $i \in \mathcal{S}^k$ in parallel}
 		\STATE $U^{k, 0}_i = U^{k}_i$
% 		\tcp*[f]{Local rounds}
		\FOR{$c = 1, \ldots, C$} %\Comment{\textit{Local rounds}}
		\STATE $\nabla_U F_i^k(U_i^{k,c}) = (I_d - U_i^{k,c}U_i^{k,c \top}) F_i^k(U_i^{k,c})$ 
		\STATE $U_{i}^{k, c+1} =R(U_i^{k, c} - \eta \nabla_U F_i^k(U_i^{k,c}))$
		\ENDFOR
		\STATE $U_i^{k+1} = U_{i}^{k, C} $
		\ENDFOR
% 		\State Aggregate all $U_i^{k+1}$ from clients
		\STATE  $CO$ updates $Z^{k+1} = \frac{1}{|\mathcal{S}^k|} \sum_{i \in \mathcal{S}^k}{} (U_i^{k+1} + \frac{1}{\rho} Y_i^k)$ %\Comment{\textit{Global update}}
		\STATE $CO$ broadcasts $Z^{k+1}$ to all clients %https://www.overleaf.com/project/60c1859569c4be9bf761503d
		  \FOR {each client $i \in \mathcal{S}^k$ in parallel} %\Comment{\textit{Local update}}
		\STATE   $Y_i^{k+1} = Y_i^k + \rho \BigP{U_i^{k+1} - Z^{k+1}} $ %\Comment{Local computation}
%		\STATE   $T_i^{k+1} = T_i^k + \rho \, h_i(U_i^{k+1})$ %\Comment{Local update}
		\ENDFOR
		\ENDFOR
	\end{algorithmic}
\end{algorithm}

% The other constraint $U^TU = I$ is based on orthogonality. 
As detailed in Alg.~\ref{algo2} (lines 5-8), projected gradient descent on the Grassmann manifold is applied in every local iteration in clients to find the update of $U_i^{k+1}$ in \eqref{E:Fopt}. Considering the definition of the Grassmann manifold in \eqref{eq:grass-constraint}, we enforce the second constraint $U_i^TU_i = I$ by using the projected gradient descent on the Grassmann manifold (Alg.~\ref{algo2}, line 8) as follows
\begin{align}
	U^{k+1}_i = R(U^k_i - \eta \nabla_U F_i^k(U^k_i)), \label{E:Gpro1}
\end{align}

Here, $\eta$ is the step size and $\nabla_U F_i^k(U^k_i)$ is the gradient of $F_i^k(U^k_i)$ at the point $U_i^k$, and the retraction function $R(\cdot)$ is the projection operation performed by QR decomposition~\cite{zhang2018grassmannian}.  The projection process for $U^{k}_i$ and $U^{k+1}_i$ is graphically illustrated in Fig.~\ref{fig:gm}. We compute the updates by projecting the Euclidean gradient onto the tangent space of the manifold using orthogonal projection $F_i^k(U^k_i)\rightarrow  \nabla_U F_i^k(U^k_i)$ (Alg.~\ref{algo2}, line 7) as follows.
\begin{equation}
	\nabla_U F_i^k(U^k_i) = (I_d - U_i^{k}U_i^{k \top})F_i^k(U^k_i). \label{E:Gpro2}
\end{equation}

Note that Equations \eqref{E:Gpro1} and \eqref{E:Gpro2} are repeatedly updated in $C$ local rounds to solve Equation~\eqref{E:Fopt} (Alg.~\ref{algo2}, lines 6-9). 

The use of Grassmannian gradients in \textsc{FedPG} enables more precise and effective steps toward the optimal solution, thus accelerating convergence. Furthermore, PCA problems often include orthogonality constraints, which are complex to handle in traditional Euclidean spaces and slow the convergence rate. \textsc{FedPG} elegantly deals with these constraints by projecting the problem onto the Grassmann manifold, where orthogonality constraints are naturally satisfied. By doing so, it eliminates the need for extra steps to manage these constraints, reducing computational overhead and facilitating faster convergence.

Moreover, \textsc{FedPG} ensures an exact solution due to its unique feature of projection onto the Grassmann manifold. This process maintains the inherent orthogonality conditions that PCA requires, effectively preserving the constraints of the original problem. Therefore, the solutions derived from \textsc{FedPG} are precise and exact as they are always in the feasible region. On the other hand, \textsc{FedPE}'s solutions are approximations. While \textsc{FedPE} works effectively in solving the problem, it doesn't inherently uphold the orthogonality conditions. The resulting solutions, therefore, might be outside the feasible region, and additional steps are necessary to bring the solution back into the feasible space. This could lead to approximation errors, and therefore the solutions are not always exact.

%The convergence of ADMM-based algorithms has been well-studied in the literature \cite{giesen2016distributed, Hong2015,Chang2016, admm_nonconvex}. To ensure that \textsc{FedPE} and \textsc{FedPG} converge to a set of stationary solutions, we provide convergence analyses in Sec.~\ref{sec:convergence_analysis}

%\blue{Justify \textsc{FedPG} can give exact solution due to projection, while \textsc{FedPE} just give approximation.}
\section{\textsc{FedPCA}: Convergence Analysis}
\label{sec:convergence_analysis}
In this section, we present a theoretical analysis to understand the convergence properties of \textsc{FedPG} and \textsc{FedPE} within the FedPCA framework. To begin with, we consider an FL environment with $N$ clients, indexed by $i=1,2,\ldots, N$, and introduce the following notations for the analyses:
%Let $i=1,2, \ldots, N$ be the indice of $i$-th client.
\begin{enumerate}
    \item $\mathcal{S}^k \subset \{1,2, \ldots, N\}$: denotes the subset of clients selected in the $k$-th iteration.
    \item $k(i)$: denotes the latest iteration at which the client $i$ has its model updated.
\end{enumerate}

%The following convergence analysis is conducted for both FedPG and FedPE. 
The orthonormality constraint in \eqref{Prob:pca_decentralized_admm} is a critical component for our proposed method. While \textsc{FedPE} handles this by introducing a surrogate linear constraint in (\ref{eqn:Hi}), \textsc{FedPG} relaxes this by leveraging projected gradient descent on Grassmann manifolds. Hence, to ensure the generality of our framework, we employ the augmented Lagrangian \eqref{eqn: lagrange} and rewrite it to adapt for the convergence analysis of both algorithms as follows.
\begin{equation}
    \label{eqn: aug_lagrangian}
    \begin{split}
	\mathcal{L}(\{U_i\}, Z, \{Y_i\}) &= \SumLim{i=1}{N} f_i(U_i) + \SumLim{i=1}{N} Y_i^\top (U_i - Z) \\
     & \qquad\qquad\qquad + \frac{\rho_i}{2} \SumLim{i=1}{N} \norm{U_i - Z}^2_F \\
     \mathrm{s.t.} \quad  U_i^\top U_i & = I.
    \end{split}
\end{equation}

In addition, we modify the global update~\eqref{ADMM_PCA_update_Z} by following:
\begin{equation}
    \label{eqn:Z_update}
    Z^{k+1} = \frac{1}{|S_k|} \sum_{i \in S_k}^{}(U_i^{k+1} +  \frac{1}{\rho_i} Y_i^k)
\end{equation}

Accordingly, the local updates for $U_i^{k+1}$ in~\eqref{ADMM_PCA_update_U_i} and $Y_i^{k+1}$ in~\eqref{ADMM_PCA_update_Y_i} at a selected client $i$ will be changed as listed below.
%Step 1:
\begin{align}
    U_i^{k+1} & = \underset{U_i}{\operatorname{\argmin}}  \BigC{f_i(U_i) + \innProd{Y_i^k, U_i - Z^k}_F \nonumber \\
    & \qquad\qquad\qquad\qquad+ \frac{\rho_i}{2} \SumLim{i=1}{N} \norm{U_i - Z^k}^2_F} \label{eqn:Ui_update} \\
     & \mathrm{s.t.} \quad  U_i^\top U_i = I \nonumber \\      
     Y_i^{k+1} & = Y_i^k + \rho_i \BigP{U_i^{k+1} - Z^{k+1}} \label{eqn:Yi_update}
\end{align}
%Step 2:
It is worth noting that this modification does not affect the performance of \textsc{FedPE} and \textsc{FedPG} since the update results will be the same after one iteration.

\textbf{Assumption.}
We first make the following assumptions:
\begin{itemize}
	\item A1: For each client's objective function $f_i, \forall i = 1, \ldots, N$ there exists a constant $L_i \geq 0$ such that for any two points $U_i, Z_i$, the gradient of $f_i$ satisfies the Lipschitz condition:  $$\left\|\nabla f_i\left(U_i\right)-\nabla {f_i}\left(Z_i\right)\right\|_F \leq L_i\left\|U_i-Z_i\right\|_F, \forall i$$
	
	\item A2: For all $i$, the penalty parameter $\rho_i$ is chosen large enough such that the $U_i$ sub-problem (\ref{eqn:Ui_update}) is strongly convex with modulus $\mu_i\left(\rho_i\right)$.

    \item A3: $\gL(x)$ is bounded from below, i.e.,
        $$\underline{\gL} \defeq \min_{X} \gL(X) > - \infty$$
\end{itemize}

\begin{remark}
 Assumptions A1 and A3 are standard in analyzing convergence properties for ADMM. Under Assumption A2, as $\rho_i$ increases, the subproblem \eqref{eqn:Ui_update} will be eventually strongly convex with respect to $U_i$. The associated strong convexity modulus $\mu_i\left(\rho_i\right)$ is a
monotonic increasing function of $\rho_i$.  
\end{remark}
%To establish a convergence guarantee  ADMM, we aim to show the following... 

To establish a convergence guarantee of \textsc{FedPE} and \textsc{FedPG}, we aim to show the following. Start with the variables $(\{ U_i^k \}, Z^k, \{ Y_i^k \})$. After $T$ rounds (i.e., after all clients have participated at least once since iteration $k$), the new variables $(\{ U_i^{k+T} \}, Z^{k+T}, \{ Y_i^{k+T} \})$ lead to decrease in the global augmented Lagrangian $\gL$. Then we show these variables will converge to a local stationary point of $\gL$. Notably, our analysis employs matrix variables, in contrast to other ADMM-based methods using vector variables~\cite{giesen2016distributed, Hong2015,Chang2016, admm_nonconvex}
%\blue{Our problem relies on matrix variables, unlike other ADMM-based algorithms using vector variables~\cite{giesen2016distributed, Hong2015,Chang2016, admm_nonconvex}}
\vspace{-15pt}
\subsection{Bound on the Successive Difference of Dual Variables}
We establish an upper bound for the successive difference of the dual variables $Y_i$ in terms of the primal variables $U_i$.
\begin{lemma}
    Suppose Assumptions A1 hold, then we have
    \begin{equation}
    \label{eqn:lemma1}
        \left\|Y_i^{k+1}-Y_i^k\right\|_F^2 \leq L_i^2\left\|U_i^{k+1}-U_i^k\right\|_F^2, \forall i, \forall k
    \end{equation}
    \label{lemma:diff_dual}
\end{lemma}
%\begin{proof}
\textit{Proof.}
In the case of $i \not\in S^k$, (\ref{eqn:lemma1}) is true because both sides of (\ref{eqn:lemma1}) are equal to zero. When $i \in S^k$, we have below equation from optimal condition for (\ref{eqn:Ui_update}).
\begin{equation}
    \label{eqn:lemma1-1}
    \nabla {f_i}\left(U_i^{k+1}\right) + Y_i^{k} + \rho_i(U_i^{k+1} - Z^{k+1}) = 0
\end{equation}
Substituting (\ref{eqn:lemma1-1}) to the dual variable update (\ref{eqn:Yi_update}), the following equation is derived.
\begin{equation}
    \label{eqn:*}
    \nabla {f_i}\left(U_i^{k+1}\right) = -Y_i^{k+1}
\end{equation}
Combining with Assumption $A1$, and noting that for any given $i$, $U_i$ and $Y_i$ are updated in the same iteration, we obtain for all $i \in {S^k} / \{0\}$ :
\begin{equation*}
\begin{split}
\left\|Y_i^{k+1}-Y_i^k\right\|_F &  = \left\|\nabla {f_i}\left(U_i^{k+1}\right)-\nabla {f_i}\left(U_i^{k(i)}\right)\right\|_F \\
&= \left\|\nabla {f_i}\left(U_i^{k+1}\right)-\nabla {f_i}\left(U_i^{k(i)}\right)\right\|_F \\
&\leq L_i\left\|U_i^{k+1}-U_i^{k(i)}\right\|_F = L_i\left\|U_i^{k+1}-U_i^k\right\|_F
\end{split}
\end{equation*}
%\end{proof}
\subsection{Bound for the Augmented Lagrangian}
Next, we show that the augmented Lagrangian \eqref{eqn: aug_lagrangian} can be decreased sufficiently and bounded below.

\begin{lemma}%[Decrease in the augmented Lagrangian]
Suppose Assumption A1, A2 and A3 hold, then the augmented Lagrangian is bounded as follows.
\begin{equation*}
\begin{split}
	& \mathcal{L}\left(\left\{U_i^{k+1}\right\}, Z^{k+1},\left\{Y_i^{k+1}\right\}\right)-\mathcal{L}\left(\left\{U_i^k\right\}, Z^k,\left\{Y_i^k\right\}\right) \\
	\leq & \sum_{i \in S^{k}}\left(\frac{L_i^2}{\rho_i}-\frac{\mu_i}{2}\right)\left\|U_i^{k+1}-U_i^k\right\|_F^2-\frac{\gamma}{2}\left\|Z^{k+1}-Z^k\right\|^2
\end{split}
\end{equation*}
\label{lemma:auglag_decrease}
\end{lemma}
%\begin{proof}
\textit{Proof.}
For each client $i \in S^k$, we can split the successive difference of the augmented Lagrangian by following.
\begin{align}
	& \mathcal{L}\left(\left\{U_i^{k+1}\right\}, Z^{k+1},\left\{Y_i^{k+1}\right\}\right)-\mathcal{L}\left(\left\{U_i^k\right\}, Z^k,\left\{Y_i^k\right\}\right) \nonumber\\
	& = \underbrace{\mathcal{L}\left(\left\{U_i^{k+1}\right\}, Z^{k+1},\left\{Y_i^{k+1}\right\}\right)-\mathcal{L}\left(\left\{U_i^{k+1}\right\}, Z^{k+1},\left\{Y_i^k\right\}\right)}_A \nonumber\\
	& +\underbrace{\mathcal{L}\left(\left\{U_i^{k+1}\right\}, Z^{k+1},\left\{Y_i^k\right\}\right)-\mathcal{L}\left(\left\{U_i^{k+1}\right\}, Z^k,\left\{Y_i^k\right\}\right)}_B \nonumber\\
	& +\underbrace{\mathcal{L}\left(\left\{U_i^{k+1}\right\}, Z^k,\left\{Y_i^k\right\}\right)-\mathcal{L}\left(\left\{U_i^k\right\}, Z^k,\left\{Y_i^k\right\}\right)}_C \label{eq:augLag:split}
\end{align}

\textbf{Bounding Term A.} The first term in \eqref{eq:augLag:split} can be bounded as follows.
\begin{align}
	& \mathcal{L}\left(\left\{U_i^{k+1}\right\}, Z^{k+1},\left\{Y_i^{k+1}\right\}\right)-\mathcal{L}\left(\left\{U_i^{k+1}\right\}, Z^{k+1},\left\{Y_i^k\right\}\right) \nonumber\\
	& =\sum_{i=1}^N\left\langle Y_i^{k+1}-Y_i^k, U_i^{k+1}-Z^{k+1}\right\rangle_F \label{eq:termA}  \\
	& \stackrel{(a)}{=} \sum_{i \in S^{k}} \frac{1}{\rho_i}\left\|Y_i^{k+1}-Y_i^k\right\|_F^2 \nonumber
\end{align}
where (a) is achieved by using (\ref{eqn:Yi_update}).

\textbf{Bounding Term B.} The second term in \eqref{eq:augLag:split} can be bounded as follows.
\begin{align}
	& \mathcal{L}\left(\left\{U_i^{k+1}\right\}, Z^{k+1}, Y^k\right)-\mathcal{L}\left(\left\{U_i^{k+1}\right\}, Z^k,\left\{Y_i^k\right\}\right) \nonumber\\
	& \stackrel{(A2)}{\leq} \sum_{i=1}^N \Big(\left\langle\nabla_{U_i} \mathcal{L}\left(\left\{U_i^{k+1}\right\}, Z^{k+1}, Y^k\right), U_i^{k+1}-U_i^k\right\rangle \nonumber \\
    & \hskip14em\relax -\frac{\mu_i}{2}\left\|U_i^{k+1}-U_i^k\right\|_F^2 \Big) \nonumber \\
	& \stackrel{(b)}{=} -\sum_{i \in S^{+}} \frac{\mu_i}{2}\left\|U_i^{k+1}-U_i^k\right\|_F^2 .\label{eq:termB}
\end{align}
where (b) is trivially true in the case of $i \not\in S^k$ as $U_i^{k+1}=U_i^k$. In the case of $i \in S^k$, the $\nabla_{U_i} \mathcal{L}\left(\left\{U_i^{k+1}\right\}, Z^{k+1}, \{Y_i^k\}\right)=0$ because of optimal condition of (\ref{eqn:Ui_update}).

\textbf{Bounding Term C.} The third term in \eqref{eq:augLag:split} can be bounded as follows.
\begin{align}
	& \mathcal{L}\left(\left\{U_i^{k+1}\right\}, Z^k, \{Y_i^k\}\right)-\mathcal{L}\left(\left\{U_i^k\right\}, Z^k, \{Y_i^k\}\right) \nonumber\\
	& \stackrel{(c)}{\leq} \sum_{i=1}^N \Big(\left\langle \nabla_{Z} \mathcal{L}\left(\left\{U_i^k\right\}, Z^{k+1}, \{Y_i^k\}\right), Z^{k+1}-Z^k\right\rangle \nonumber\\
    & \hskip14em\relax-\frac{\gamma}{2}\left\|Z^{k+1}-Z^k\right\|_F^2\Big) \nonumber\\
	& \stackrel{(d)}{=} \sum_{i \in S^{k}}-\frac{\gamma}{2}\left\|Z^{k+1}-Z^k\right\|_F^2 . \label{eq:termC}
\end{align}
where in (c) we used the fact that $\mathcal{L}\left(\left\{U_i\right\}, Z, \{Y_i\}\right)$ is strongly convex w.r.t. $Z$. 

For (d), it is trivially true if $i \not \in S^k$ as $U_i^{k+1}=U_i^k$. When $i\in S^k$, the $\nabla_Z \mathcal{L}\left(\left\{U_i^k\right\}, Z^{k+1}, \{Y_i^k\}\right)=0$ because of the optimal condition of (\ref{eqn:Ui_update}).

\textbf{Sufficient decrease.} Combining the inequalities \Cref{eq:termA}, \Cref{eq:termB} and \Cref{eq:termC} together, we have 
\begin{align}
    & {A+B+C} \nonumber\\ 
    & = \mathcal{L}\left(\left\{U_i^{k+1}\right\}, Z^{k+1},\left\{Y_i^{k+1}\right\}\right)-\mathcal{L}\left(\left\{U_i^k\right\}, Z^k,\left\{Y_i^k\right\}\right) \label{eq:suff_dec} \\ & \leq \sum_{i \in S^k}-\frac{\gamma}{2}\left\|Z^{k+1}-Z^k\right\|_F^2
      +\left(\frac{L_i^2}{\rho_i}-\frac{\mu_i}{2}\right)\left\|U_i^{k+1}-U_i^k\right\|_F^2   \nonumber
\end{align}
%\hskip10em\relax
This implies that the value of the augmented Lagrangian will always decrease if the following condition is satisfied:
\begin{equation}
    \rho_i \mu_i \geq 2 L_i^2
    \label{eq:condition}
\end{equation}
\textbf{Lower bound for the augmented Lagrangian.} We continue showing that $\mathcal{L}\left(\left\{U_i^{k}\right\}, Z^{k}, \{Y_i^{k}\}\right)$ convergence, i.e., $\mathcal{L}$ is bounded below. 

\begin{equation*}
\begin{split}
	& \mathcal{L}\left(\left\{U_i^{k+1}\right\}, Z^{k+1}, \{Y_i^{k+1}\}\right) \\
    & =  \sum_{i=1}^N\Big(f_i\left(U_i^{k+1}\right) +\left\langle Y_i^{k+1}, U_i^{k+1}-Z^{k+1}\right\rangle \\
    &  \hskip14em\relax + \frac{\rho_i}{2}\left\|U_i^{k+1}-Z^{k+1}\right\|_F^2\Big)\\
    & \stackrel{(e)}{=}\sum_{i=1}^N\Big(f_i\left(U_i^{k+1}\right) +\left\langle\nabla f\left(U_i^{k+1}\right), Z^{k+1}-U_i^{k+1}\right\rangle \\
    & \hskip14em\relax +\frac{\rho_i}{2}\left\|U_i^{k+1}-Z^{k+1}\right\|_F^2\Big) \\
    & \stackrel{(f)}{\geq} \sum_{i=1}^N f_i\left(Z^{k+1}\right)=\underline{\gL}\left(Z^{k+1}\right)
\end{split}
\end{equation*}
where (e) can be explained as following:

(e.1) If $i \in S^{k}$, based on (\ref{eqn:*}):
\[\left\langle Y_i^{k+1}, U^{k+1}-Z^{k+1}\right\rangle=\left\langle\nabla_f\left(U_i^{k+1}\right), Z^{k+1}-U^{k+1}\right\rangle\]
(e.2) If $i \notin S^{k}$, $Y_i^{k+1}=Y_i^k=Y_i^{k(i)}$, $U_i^{k+1}=U_i^k=U^{k(i)}$
\begin{equation*}
    \begin{split}
	&\left\langle Y_i^{k+1}, U_i^{k+1}-Z^{k+1}\right\rangle \\
    & =\left\langle\nabla f\left(U_i^{k(i)}\right), Z^{k+1}-U_i^{k(i)}\right\rangle \\
	&=\left\langle\nabla f\left(U_i^{k+1}\right), Z^{k+1}-U_i^{k+1}\right\rangle
\end{split}
\end{equation*}

From combining (e.1) and (e.2), we deduce that (e) holds true. Consequently, due to the Lipschitz continuity of the gradient of $f_i$, (f) is also true.
%\end{proof}
\begin{remark}
    \Cref{lemma:auglag_decrease} shows that the augmented Lagrangian decreases by a sufficient amount after each iteration. In practice, however, the penalty parameter  $\rho$ is difficult to estimate. Hence, for our experiments, we fine-tune $\rho$ from a predefined range of values to select the best one.
\end{remark}

\subsection{Convergence of the augmented Lagrangian}
\label{ap:thrm_proof:auglag_convergence}
Here, we combine Lemma \eqref{lemma:diff_dual} and \eqref{lemma:auglag_decrease} to establish the following theorem.

\begin{theorem}
    Suppose the following are true:
    \begin{itemize}
        \item After $T$ iterations, all clients have participated in training at least once.
        \item The global loss function $\mathcal{L}(Z^{k}, \{ U_i^{k} \}, \{ Y_i^{k} \})$ is bounded below by a finite quantity $\underline{\gL}$.
        \item The hyperparameter $\rho_i$ for the augmented Lagrangian is chosen such that $\rho_i \mu_i \geq 2 L_i^2$.
    \end{itemize}
    Then the augmented Lagrangian $\gL$ will monotonically decrease and is convergent to a quantity of at least $\underline{\gL}$. Further, for all $i=1, \dots, N$ we have $\lim_{k \rightarrow \infty} \norm{Z^{k+T} - U_i^{k+T}} = 0$.
    \label{appn:admm:auglag_convergence}
\end{theorem}

%\begin{proof}
\textit{Proof.}
\Cref{appn:admm:auglag_convergence} implies that the augmented Lagrangian decreases by a non-negative amount after every $T$ global iterations. Given the fact that every client will be updated at least once in the interval $[k, k + T]$, we conclude that the limit $\lim_{k \rightarrow \infty} \mathcal{L}(Z^{k+T}, \{ U_i^{k+T} \}, \{ Y_i^{k+T} \})$ exists and is at least $\underline{\gL}$.

Now we prove the second statement. From \Cref{eq:suff_dec} and the fact that $\gL$ converges, we conclude that as $k \rightarrow \infty$, 
\begin{align*}
    \norm{Z^{k+T} - Z^{k}} \rightarrow 0, \quad \norm{U_i^{k+T} - U_i^{k}} \rightarrow 0.
\end{align*}
Combining this with \Cref{lemma:diff_dual}, we have $\norm{Y_i^{k+1} - Y_i^{k}} \rightarrow 0$. Based on the definition of $Y_i^{t+1}$, this implies that $$\norm{U_i^{k+1} - Z^{k+1}} \rightarrow 0$$ 
%\end{proof}
\begin{remark}
\Cref{appn:admm:auglag_convergence} establishes that the sequence of primal and dual variables updated after each $T$ global iterations of \Cref{algo1} and \Cref{algo2} converges. We further have $\norm{U_i^{k+1} - Z^{k+1}} \rightarrow 0$, i.e., the consensus constraint is satisfied.
\end{remark}

\subsection{Convergence to stationary point}
\label{ap:thrm_proof:auglag_stationary}
In the below theorem we present another guarantee that the limit of the sequence $(Z^{k}, \{ U_i^{k} \}, \{ Y_i^{k} \})$ is the a stationary solution for Problem \Cref{Prob:pca_decentralized_admm_nonlinear}.
\begin{theorem} 
    Suppose the assumptions in \Cref{appn:admm:auglag_convergence} hold. Then the limit point $(Z^{*}, \{ U_i^{*} \}, \{ Y_i^{*} \})$ of the sequence $(Z^{k}, \{ U_i^{k} \}, \{ Y_i^{k} \})$ is a stationary solution to Problem \Cref{Prob:pca_decentralized_admm_nonlinear}. That is, for all $i$,
    \begin{align*}
        \nabla f_i(Z^*) + Y_i^* = 0 \text{ and } Z^* = U_i^*.
    \end{align*}
\label{appn:admm:auglag_convergence_stationary}
\end{theorem}
%\vspace{-20pt}
\textit{Proof.} See the proof to \cite[Theorem 2.4]{admm_nonconvex}. Here we make some remarks. First, because \textsc{FedPE} and \textsc{FedPG} allow for client sampling, the proof relies on an assumption that all clients will participate after some fixed number of rounds (in particular, $T$ rounds). Second, since local objectives are smooth (possibly non-convex), the primal variables $U_i$ are updated using first-order methods like SGD or projected GD. Finally, the proof aims to show that after every $T$ rounds, we obtain a sufficient decrease in the augmented Lagrangian; coupled with the fact that the augmented Lagrangian is bounded below, the sequence of variables converges to a stationary point of $\gL$.
%\end{proof}
%\textcolor{blue}{
\subsection{Convergence rate of the augmented Lagrangian}
We first define gradient of the Augmented Lagrangian as:
\begin{align}
    \hat{\nabla} \mathcal{L}\left(\left\{U_i\right\}, Z, Y\right):=\left[\begin{array}{c}\nabla_{U_1} \mathcal{L}\left(\left\{U_i\right\}, Z, Y\right) \\ \nabla_{U_2} \mathcal{L}\left(\left\{U_i\right\}, Z, Y\right) \\ \vdots \\ \nabla_{U_N} \mathcal{L}\left(\left\{U_i\right\}, Z, Y\right)\end{array}\right] \nonumber
\end{align}
We aim to use the below quantity to establish the progress of the proposed methods:
\begin{equation}
\begin{split}
  &P(\{U^{k}_{i}\}, Z^{k}, Y^{k}_{i}):= \sum_{i=1}^{N}||U^{k}_{i} - Z^{k}||^2 \\
  &+ ||\hat{\nabla}\mathcal{L}(\{U^{k}_{i}\}, Z^{k}, Y^{k}_{i})||^2 \nonumber \\
\end{split}
\end{equation}
It can be verified that if $P(\{U^{k}_{i}\}, Z^{k}, Y^{k}_{i}) \rightarrow 0$, then a stationary solution is obtained, which $U_i$ converges to $Z$ and gradient for Augmented Lagrangian of each client $i$ is equal to zero.
\begin{theorem}
    Suppose that Assumptions A.1, A.2, A.3 and Theorem \ref{appn:admm:auglag_convergence} are satisfied. Let $T(\epsilon) $ denote an iteration index in which the following inequality is achieved:
    \begin{equation}
        T(\epsilon):=\min(\{k|P(\{U^{k}_{i}\}, Z^{k}, Y^{k}_{i})\} \leq \epsilon, k\ge0) \nonumber
    \end{equation}
    for some $\epsilon > 0$. Then there exists some constant $\beta>0$ such that
    \begin{equation}
    T(\epsilon) \leq   \frac{\beta(\mathcal{L}(\{U^{1}_{i}\}, Z^{1}, Y^{1}_{i}) - \underline{\mathcal{L}})}{\epsilon} \nonumber
    \end{equation}
    where $\underline{\mathcal{L}}$ is defined by Assumption A.3.
    \label{theo:convergence}
\end{theorem}
%}
%\textcolor{blue}{
%\begin{proof}
%\textcolor{blue}{
\textit{Proof.} Based on the optimal condition of $U_i$ \eqref{ADMM_PCA_update_U_i}:
\begin{equation}
    \nabla f_i(U^{k+1}_{i}) + Y^{k}_i + \rho_i(U^{k+1}_i - Z^{k+1}) = 0
    \label{r3:optimal_condition_Ui}
\end{equation}
%}
%\textcolor{blue}{
By definition of $\mathcal{L}$, we can have:
\begin{equation}
    \begin{split}
        &||\nabla_{U_i} \mathcal{L}(\{U^{k}_{i}\}, Z^{k}, Y^{k}_{i})|| = ||\nabla f_i(U^{k}_{i}) + Y^{k}_i + \rho_i(U^{k}_i - Z^{k})|| \\ 
        &\stackrel{(g)}{=} ||\nabla f_i(U^{k}_{i}) + Y^{k}_i + \rho_i(U^{k}_i - Z^{k}) \\
        &- (\nabla f_i(U^{k+1}_{i}) + Y^{k}_i + \rho_i(U^{k+1}_i - Z^{k+1}))|| \\
        &\leq ||\nabla f_i(U^{k}_{i}) - \nabla f_i(U^{k+1}_{i})|| + \rho_i||U^{k+1}_i - U^{k}_i||\\
        % + ||Y^{k+1}_i - Y^{k}_i|| \\
        &+ \rho_i||Z^{k+1} - Z^{k}||  \\
        &\stackrel{(h)}{\leq} L_i||U^{k+1}_{i} - U^{k}_{i}|| + \rho_i||U^{k+1}_i - U^{k}_i|| + \rho_i||Z^{k+1} - Z^{k}||  \\
        &=(L_i + \rho_i)||U^{k+1}_{i} - U^{k}_{i}||+ \rho_i||Z^{k+1} - Z^{k}||
    \end{split}
    \label{convergence_rate_1}
\end{equation}
Where (g) is obtained by optimal condition of $U_i$ in \eqref{r3:optimal_condition_Ui}. To derive (h), the first term is bounded by assumption $A.1$ and the second term is bounded by Lemma \ref{lemma:diff_dual}.
%}
%\textcolor{blue}{
Based on \eqref{convergence_rate_1}, we have the following:
\begin{equation}
\begin{split}
    &||\hat{\nabla}\mathcal{L}(\{U^{k}_{i}\}, Z^{k}, Y^{k}_{i})|| \leq \sum_{i=1}^{N}(L_i + \rho_i)||U^{k+1}_{i} - U^{k}_{i}||\\
    &+ \sum_{i=1}^{N}\rho_i||Z^{k+1} - Z^{k}||  
\end{split}
\end{equation}
%}
%\textcolor{blue}{
By taking $\sigma_1 = \max \{\sum_{i=1}^{N}\rho_i, L_1 + \rho_1, ..., L_N + \rho_N\}$, it is apparent that $\sigma_1 > 0$. We can have following:
%}
%\textcolor{blue}{
\begin{equation}
||\hat{\nabla}\mathcal{L}(\{U^{k}_{i}\}, Z^{k}, Y^{k}_{i})|| \leq \sigma_1(||Z^{k+1} - Z^{k}|| + \sum_{i=1}^{N}||U^{k+1}_{i} - U^{k}_{i}||)  
\label{r3:convergence_rate_3}
\end{equation}
%}
%\textcolor{blue}{
From \eqref{ADMM_PCA_update_Y_i}, we have
\begin{equation}
\begin{split}
    &\sum_{i=1}^{N}||U^{k+1}_{i} - Z^{k+1}|| = \sum_{i=1}^{N}\frac{1}{\rho_i}||Y^{k+1}_{i} - Y^{k}_{i}|| \\
    &\leq \sum_{i=1}^{N}\frac{L_i}{\rho_i}||U^{k+1}_{i} - U^{k}_{i}||
    \label{r3:convergence_rate_4}
\end{split}
\end{equation}
\eqref{r3:convergence_rate_3} and \eqref{r3:convergence_rate_4} imply for some $\sigma_3 \geq 0$:
\begin{equation}
\begin{split}
        &\sum_{i=1}^{N}||U^{k}_{i} - Z^{k}||^2 + ||\hat{\nabla}\mathcal{L}(\{U^{k}_{i}\}, Z^{k}, Y^{k}_{i})||^2 \\
        &\leq \sigma_3( ||Z^{k+1} - Z^{k}||^2 + \sum_{i=1}^{N}||U^{k+1}_{i} - U^{k}_{i}||^2)\\
    \label{r3:convergence_rate_5}
\end{split}
\end{equation}
It is noteworthy that we have changed from $||U^{k+1}_{i} - Z^{k+1}||^2$ to $||U^{k}_{i} - Z^{k}||^2$. This is because we have proved that $||U^{k+1}_{i} - Z^{k+1}|| \rightarrow 0$, which means the gap between these two terms $||U^{k+1}_{i} - Z^{k+1}||^2$ and $||U^{k}_{i} - Z^{k}||^2$ are finite. Therefore, we always can find a finite number $\sigma_3\ge0$ to satisfy $||U^{k}_{i} - Z^{k}||^2 \leq \sum_{i=1}^{N}\sigma_3||U^{k+1}_{i} - U^{k}_{i}||^2$.
From lemma \ref{lemma:auglag_decrease}, there exists a constant $\sigma_2 = min\{\frac{L^{2}_{i}}{\rho_i} - \frac{\mu_i}{2}, \mu\}_{i=1}^N$ such that:
\begin{equation}
\begin{split}
    &\mathcal{L}(\{U^{k}_{i}\}, Z^{k}, Y^{k}_{i}) - \mathcal{L}(\{U^{k+1}_{i}\}, Z^{k+1}, Y^{k+1}_{i}) \\
    &\ge \sigma_2(||Z^{k+1} - Z^{k}||^2 + \sum_{i=1}^{N}||U^{k+1}_{i} - U^{k}_{i}||^2)
    \label{r3:convergence_rate_6}
\end{split}
\end{equation}
Combining \eqref{r3:convergence_rate_5} and \eqref{r3:convergence_rate_6} we have:
\begin{equation}
\begin{split}
    &\sum_{i=1}^{N}||U^{k}_{i} - Z^{k}||^2 + ||\hat{\nabla}\mathcal{L}(\{U^{k}_{i}\}, Z^{k}, Y^{k}_{i})||^2 \\
    &\leq \frac{\sigma_3}{\sigma_2}(\mathcal{L}(\{U^{k}_{i}\}, Z^{k}, Y^{k}_{i}) - \mathcal{L}(\{U^{k+1}_{i}\}, Z^{k+1}, Y^{k+1}_{i}))
\end{split}
\end{equation}
Summing both sides of the above inequality over $k = 1, ..., T$. We have the following:
\begin{equation}
\begin{split}
    &\sum_{k=1}^{T}(\sum_{i=1}^{N}||U^{k}_{i} - Z^{k}||^2 + ||\hat{\nabla}\mathcal{L}(\{U^{k}_{i}\}, Z^{k}, Y^{k}_{i})||^2) \\
    &\leq \frac{\sigma_3}{\sigma_2}(\mathcal{L}(\{U^{1}_{i}\}, Z^{1}, Y^{1}_{i}) - \mathcal{L}(\{U^{T+1}_{i}\}, Z^{T+1}, Y^{T+1}_{i})) \\
    &\leq \frac{\sigma_3}{\sigma_2}(\mathcal{L}(\{U^{1}_{i}\}, Z^{1}, Y^{1}_{i}) - \underline{\mathcal{L}})
    \label{r3:convergence_rate_7}
\end{split}
\end{equation}
The last inequality uses the fact that $L(\{U^{T+1}_{i}\}, Z^{T+1}, Y^{T+1}_{i})$ is decreased and lower bounded by $\underline{\mathcal{L}}$ by $A.3$.
% We introduce a new variable:
% \begin{equation}
% \begin{split}
%   &P(\{U^{k}_{i}\}, Z^{k}, Y^{k}_{i}):= \sum_{i=1}^{N}||U^{k}_{i} - Z^{k}||^2 + ||\hat{\nabla}\mathcal{L}(\{U^{k}_{i}\}, Z^{k}, Y^{k}_{i})||^2 \nonumber \\
% \end{split}
% \end{equation}
% It can be verified that if $P(\{U^{k}_{i}\}, Z^{k}, Y^{k}_{i}) \rightarrow 0$, then a stationary solution is obtained, which $U_i$ converges to $Z$ and gradient for Augmented Lagrangian of each client $i$ is equal to zero. 
By putting $T(\epsilon):=\min\{k|P(\{U^{k}_{i}\}, Z^{k}, Y^{k}_{i}) \leq \epsilon, k\ge0\}$, we can rewrite \eqref{r3:convergence_rate_7} as follows.
\begin{equation}
    T(\epsilon) \epsilon \leq \frac{\sigma_3}{\sigma_2}(\mathcal{L}(\{U^{1}_{i}\}, Z^{1}, Y^{1}_{i}) - \underline{\mathcal{L}})
\end{equation}
or
\begin{equation}
    T(\epsilon) \leq \frac{\sigma_3}{\sigma_2} \frac{(\mathcal{L}(\{U^{1}_{i}\}, Z^{1}, Y^{1}_{i}) - \underline{\mathcal{L}})}{\epsilon}
    \label{r3:convergence_rate_10}
\end{equation}
This means that for a given $\epsilon$, the proposed algorithm converges with rate of $O(\frac{1}{T(\epsilon)})$. %In the worst case of the sub-sampling scheme, the client $i$, whose has the largest distance w.r.t $\underline{\mathcal{L}}$ is selected only once every $T$ rounds. We can think $r=nT$, where $n$ is number of times the client acquiring largest distance w.r.t $\underline{\mathcal{L}}$ is selected, this guarantees convergence the proposed algorithms with the rate of $O(\frac{1}{nT})$ $\sim$ $O(\frac{1}{T(\epsilon)})$ because $T(\epsilon) \geq nT$ when $r \rightarrow \infty$. 
We note that the use of $\sum_i^N$ in the above proof does not mean that we require participation from all clients. This is used to find an upper bound. Let $\beta=\frac{\sigma_3}{\sigma_2}$, theorem \ref{theo:convergence} is achieved.
%}
%\end{proof}
%}
%\textcolor{blue}{
\subsection{Convergence rates of the proposed methods}
\begin{theorem}
Let $T(\epsilon)$ be defined as in Theorem \ref{theo:convergence}. Suppose the local update \eqref{ADMM_PCA_update_U_i} in FedPE is performed using gradient descent with $C$ iterations. Then, the convergence rates for FedPE and FedPG are  $O\left(\frac{1}{CT(\epsilon)}\right)$ and $O\left(\frac{1}{C^2T(\epsilon)}\right)$, respectively.
\end{theorem}
%\begin{proof}
\textit{Proof.}
    Based on Theorem \ref{theo:convergence}, the general convergence rates of $O(\frac{1}{T(\epsilon)})$ are proved for both \textsc{FedPE} and \textsc{FedPG}. While the learning rate of solving problem \eqref{E:Fopt} can be accelerated by using gradient descent on Grassmann manifold with convergence rate of $O(1/C^2)$ \cite[Theorem 1]{lin2020accelerated}, convergence rate for gradient descent on Euclidean space to solve \eqref{ADMM_PCA_update_U_i} is $O(1/C)$ for strongly convex functions \cite{bubeck2015convex}. Therefore, the convergence rates of \textsc{FedPE} and \textsc{FedPG} can be estimated as $O(\frac{1}{CT(\epsilon)})$ and $O(\frac{1}{C^2T(\epsilon)})$ respectively.
%\end{proof}
%}

\section{Experimental Results}
\label{sec:experiment}
In this section, we first present the datasets, model settings and measures for performance assessment. Subsequently, to illustrate the efficacy of the proposed methods, we contrast \textsc{FedPE} and \textsc{FedPG} with various benchmark strategies within the context of an IDS deployment in IoT networks.

\subsection{Experimental settings}

\begin{figure}[h]
	\centering
 	\begin{subfigure}{.49\linewidth}
		\centering
		\includegraphics[width=\textwidth]{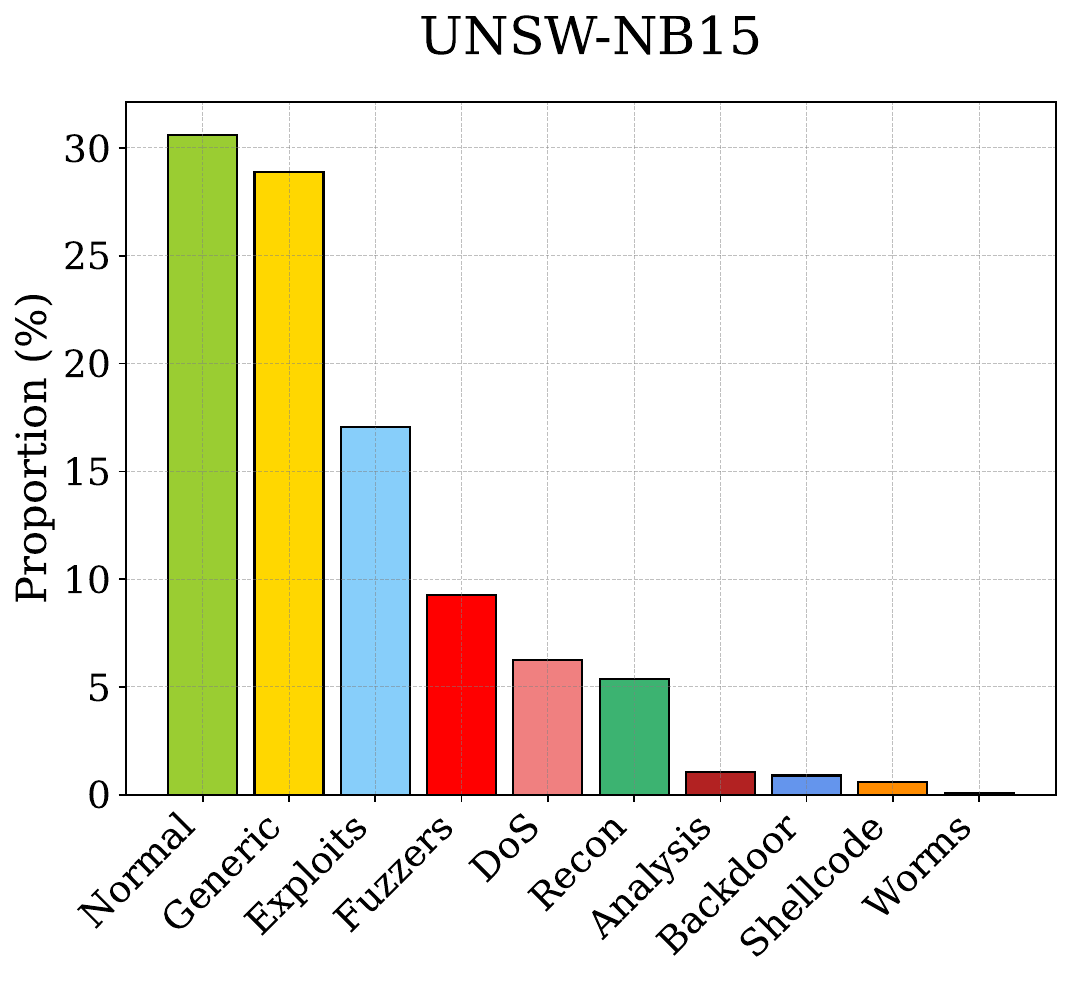}
        %\vspace{-15pt}
		\caption{}
		\label{fig:unsw}
	\end{subfigure}%
	\begin{subfigure}{.49\linewidth}
		\centering
		\includegraphics[width=\textwidth]{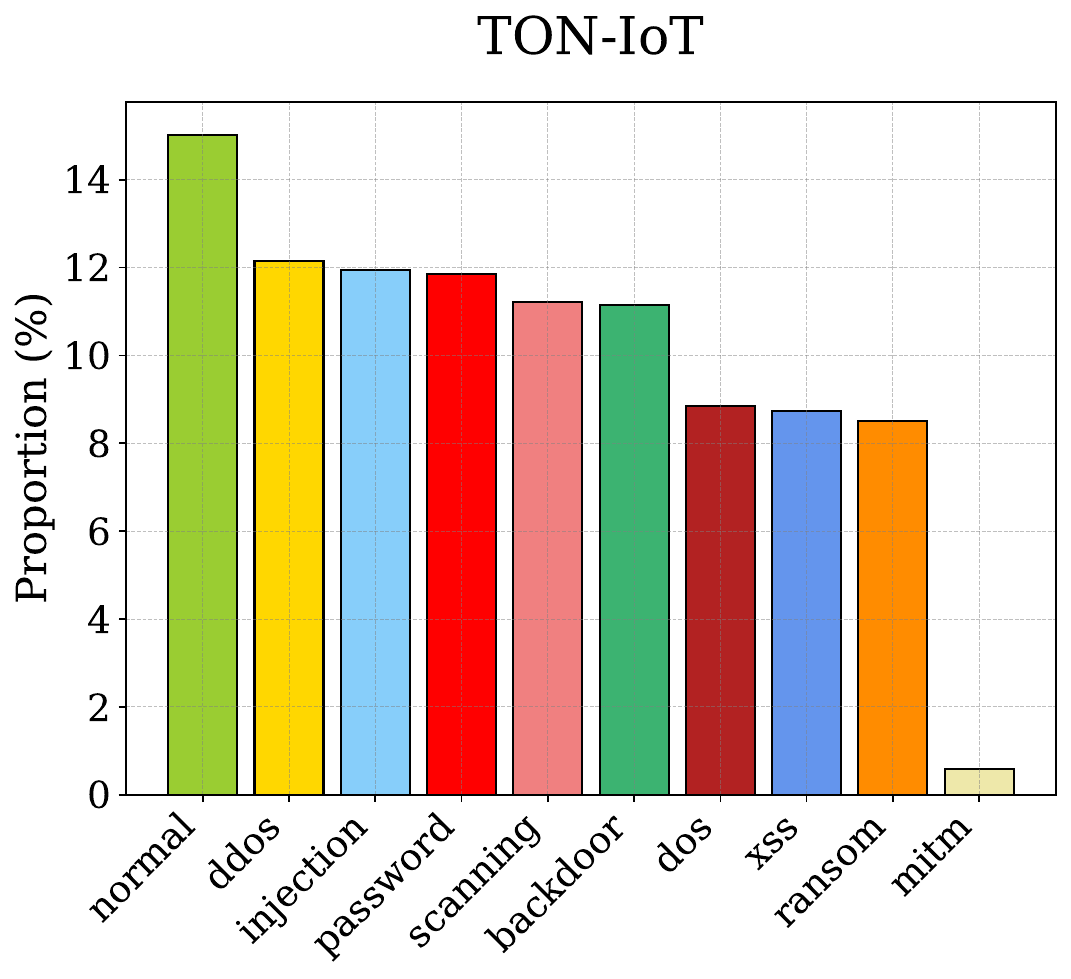}
        %\vspace{-15pt}
		\caption{}
		\label{fig:ton}
	\end{subfigure}
	%\label{fig:repochs}
	\caption{Class distribution of test sets for UNSW-NB15 and TON-IoT}
    \label{fig:datasets}
	%\vspace{-15pt}
\end{figure}

\comment{
\begin{table}[ht]
\centering
\caption{Configuration of AutoEncoder and BiGAN Models}
\renewcommand{\arraystretch}{1.15} % For row height
\scalebox{0.9}{
\begin{tabular}{|c|c|c|c|c|c|}
\hline
\textbf{Parameter} & \multicolumn{2}{c|}{\textbf{AutoEncoder}} & \multicolumn{3}{c|}{\textbf{BiGAN}} \\
\cline{2-6}
 & \textbf{Enc} & \textbf{Dec} & \textbf{$\mathcal{D}$} & \textbf{$\mathcal{G}$} & \textbf{Enc} \\
\hline
\hline
No. Layers & 3 & 3 & 3 & 3 & 3 \\
\hline
No. Hidden Units & 32 & 32 & 32 & 32 & 32 \\
\hline
Latent Dim & 2 & - & - & 2 & 2 \\
\hline
Activation & ReLU & ReLU & ReLU & ReLU & ReLU \\
\hline
Learning Rate & 0.1 & 0.1 & 0.1 & 0.1 & 0.1 \\
\hline
Local epochs & 30 & 30 & 30 & 30 & 30 \\
\hline
Batch Size & 64 & 64 & 64 & 64 & 64 \\
\hline
\end{tabular}}
\label{tab:ae_bigan}
\end{table}
}

In our research, all experiments are conducted in a coding environment powered by an AMD Ryzen 3970X Processor with 64 cores and 256GB of RAM. Additionally, four NVIDIA GeForce RTX 3090 GPUs are employed to accelerate the training process. For software, we rely on Python 3 as our primary language, supplemented by advanced machine learning and optimization frameworks such as PyTorch~\cite{pytorch} and Scikit-Learn~\cite{sklearn}. Our code can be found at: \href{https://github.com/dual-grp/FedPCA_AnomalyDetection}{https://github.com/dual-grp/FedPCA\_AnomalyDetection}.

\subsubsection{Datasets and Metrics} In this study, we employ the UNSW-NB15 \cite{unsw} and TON-IoT Network \cite{ton} datasets. The former, a benchmark in network intrusion detection system (NIDS) research, encompasses a myriad of network traffic scenarios, with various attack types interlaced with normal activities commonly found in real-world network (internet) environments. The latter, TON-IoT, includes heterogeneous data sourced from Telemetry datasets of IoT and IIoT sensors, amassed through parallel processing to capture a spectrum of normal and cyber-attack events. Following preprocessing, from the UNSW-NB15 dataset, we utilized a training set of 56,000 normal samples and a test set comprising 20,000 normal and 45,332 abnormal samples, all with 39 numerical features. From the TON-IoT dataset, our training set consists of 114,956 normal samples, while the test set includes 10,000 normal samples and 56,557 abnormal samples, all with 49 numerical features. The class distribution of the test sets for both datasets is illustrated in Fig.\ref{fig:unsw} and Fig.\ref{fig:ton} respectively. These carefully selected and prepared datasets serve as a solid foundation for the evaluation of our proposed approach.

In the setting of a host-based ML-IDS deployment, each IoT device functions as a localized ML-IDS, using its unique data to develop an anomaly detection model. This approach reflects the inherent diversity of traffic distribution across disparate clients, and to mimic this realistically, the comprehensive training set is equally subdivided into 100 individual client datasets, each representing non-identically distributed (non-i.i.d.) local data. This subdivision is achieved by grouping records based on the number of connections that contain the same service (\textit{ct\_srv\_src}) for UNSW-NB15 and the range of source ports (\textit{srcports}) for TON-IoT.

\subsubsection{Baselines} In order to highlight the efficacy of \textsc{FedPE} and \textsc{FedPG} in  IoT network anomaly detection, we conduct a comprehensive comparative study against a variety of unsupervised techniques including: 
\begin{enumerate}[a)]
    \item \textit{Self-learning PCA}: a stand-alone method where each client individually learns its PCA matrix using only its local dataset. By identifying the principal patterns within this dataset, anomalies are detected when there's a substantial discrepancy between the original data and its PCA-reconstructed version.
    \item \textit{AutoEncoder (AE)}: an unsupervised learning method constructed of an encoder and a decoder in the form of neural networks. AE facilitates anomaly detection by compressing input data during encoding and then attempting to reconstruct it. Anomalies are identified when there's a notable difference between the input data and its reconstructed version, suggesting the input doesn't conform to typical patterns within the dataset \cite{ae1}.
    \item \textit{Bidirectional Generative Adversarial Network (BiGAN)}: a generative model composed of a generator, a discriminator and an encoder. During BiGAN's adversarial training, the encoder maps real data into latent representations. The generator then tries to recreate data from these representations, aiming to closely match the original input. Concurrently, the discriminator distinguishes between pairs of real data and their latent encodings, and the output from the generator with its corresponding latent representations. When trained solely on normal data, BiGAN can detect anomalies by noting high reconstruction errors, which indicate deviations from the learned representations of normal data \cite{bigan,bigan1}.
\end{enumerate}
%Their unique approaches enables the capture of richer data representations, posing a substantial benchmark for \textsc{FedPG} and \textsc{FedPE}. 
We subsequently employ the de-facto FedAvg\cite{fedavg} to ensure consistency in the training of AE and BiGAN within federated settings. In line with this, we establish  FedAE and FedBiGAN -- which utilize two-layer neural networks for their components, and FedAE-2 and FedBiGAN-2 -- which incorporate four-layer neural networks for more complex non-linear structures. For Self-learning PCA, we train a distinct PCA matrix tailored for each client's dataset and then assess the average performance across all clients on detecting anomalies on the test set. The selection of these baseline models, widely acclaimed in the field and each offering distinct methodologies for anomaly detection, ensures a robust evaluation of our methods against diverse unsupervised paradigms.

\subsubsection{Performance Metrics}
In this work, we are primarily concerned with identifying anomalies in network traffic. Hence, the main metrics employed are:

\begin{enumerate}
    %\item \textbf{Mean Squared Error (MSE)}: Commonly employed with neural network-driven anomaly detectors, MSE delineates the average squared discrepancy between the predicted and true values.
    \item \textbf{Accuracy}: calculates the ratio of correctly identified both normal and anomalies to the total instances in the dataset, offering a general perspective on the model performance.
    \item \textbf{Precision}: represents the ratio of true anomalies to all detections labeled as anomalies, assessing the exactness of the identified anomalies.
    \item \textbf{Recall}:  quantifies the proportion of actual anomalies that are correctly flagged by the detection system, assessing the ability of a model to capture all potential threats. 
    \item \textbf{F1-Score}: calculates the harmonic mean of precision and recall, underscoring the balance between the detection of true anomalies and the risk of false alarms.
    \item \textbf{False Negative Rate (FNR)}: Represents the fraction of actual anomalies that are incorrectly flagged as normal.
    %\item \textbf{True Positive Rate (TPR)}: Often referred to as sensitivity or recall, it quantifies the proportion of true anomalies that are correctly detected.  
    %\item \textbf{False Positive Rate (FPR)}: Represents the fraction of benign instances that are inaccurately labeled as anomalous.
    \item \textbf{Area Under the Receiver Operating Characteristic curve (AUC-ROC)}: A graphical representation plotting True Positive Rate (TPR) against False Positive Rate (FPR) across varying threshold settings, furnishing a comprehensive performance overview.
    \item \textbf{Precision-Recall (PR) curves}: Serving as a visual instrument to gauge the relationship between precision and recall, this becomes vital, particularly in imbalanced datasets where anomalies are few.
\end{enumerate}
%These metrics, in tandem, constitute a robust framework for evaluating the proficiency of anomaly detection mechanisms.

\subsubsection{Training Details}  To provide a practical test for our proposed methods, we train our federated PCA models on each client's dataset, without inter-client data sharing. The comprehensive 
testing set, inclusive of unknown attacks, is applied uniformly to all methods. Preceding training, all features undergo normalization through a z-score function, utilizing the standard deviation and mean values calculated from the corresponding training set. The following experiments have hyperparameters fine-tuned via a grid search to ensure optimal test performance. Furthermore, each communication round includes a randomly selected subset of clients, constituting 10\% of the total count, replicating real-world scenarios where not all devices might be available for communication simultaneously.

\subsection{Experimental Results}

\subsubsection{Efficiency of Federated PCA on Grassmann Manifold}
\begin{figure*}[t]
	\centering
 	\begin{subfigure}{.24\textwidth}
		\centering
		\includegraphics[width=\textwidth]{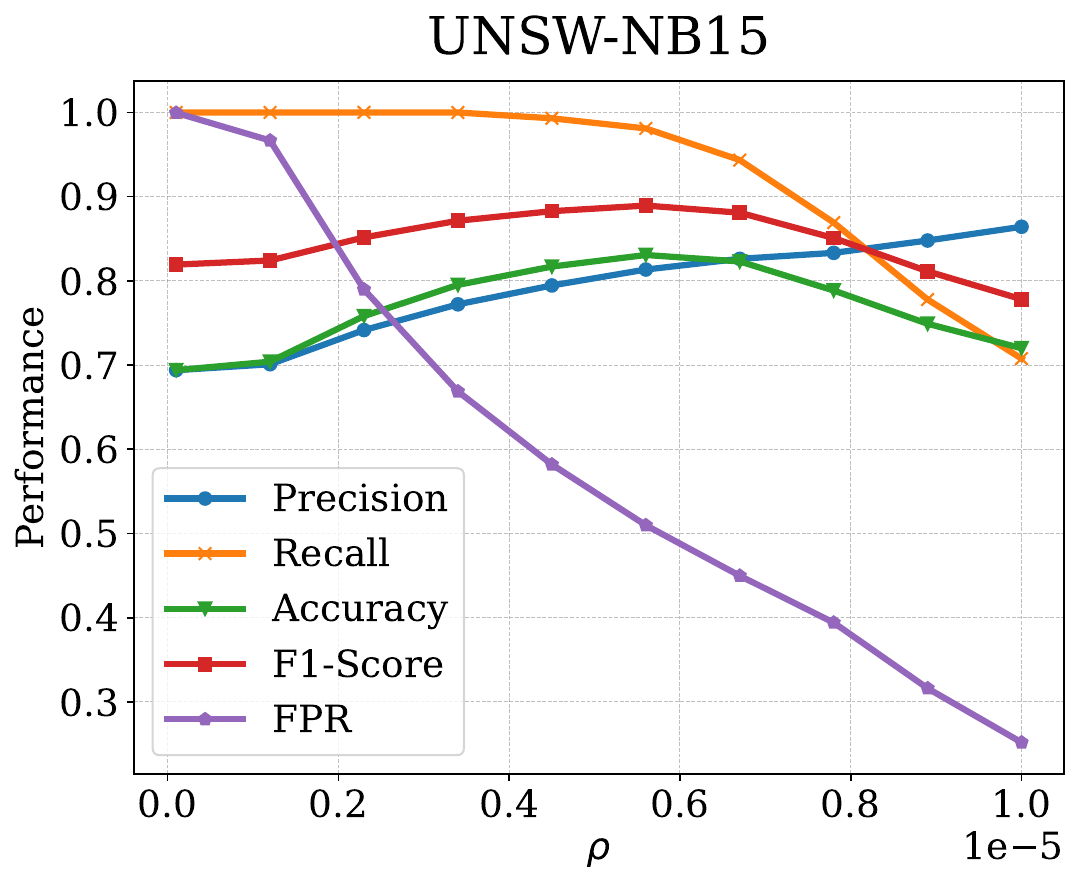}
        \vspace{-15pt}
				\caption{}
		\label{fig:finetuning5}
	\end{subfigure}%
	\begin{subfigure}{.24\textwidth}
		\centering
		\includegraphics[width=\textwidth]{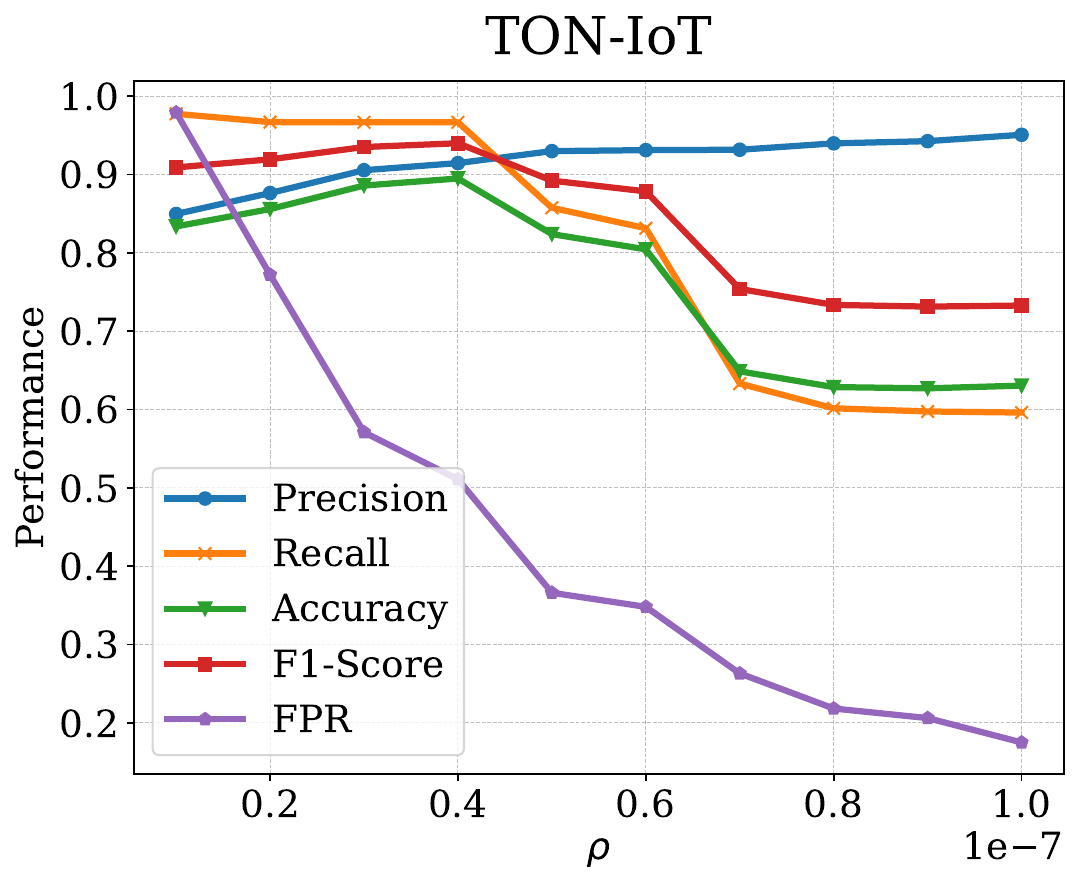}
        \vspace{-15pt}
            \caption{}
		\label{fig:finetuning6}
	\end{subfigure}
 	\begin{subfigure}{.24\textwidth}
		\centering
		\includegraphics[width=\textwidth]{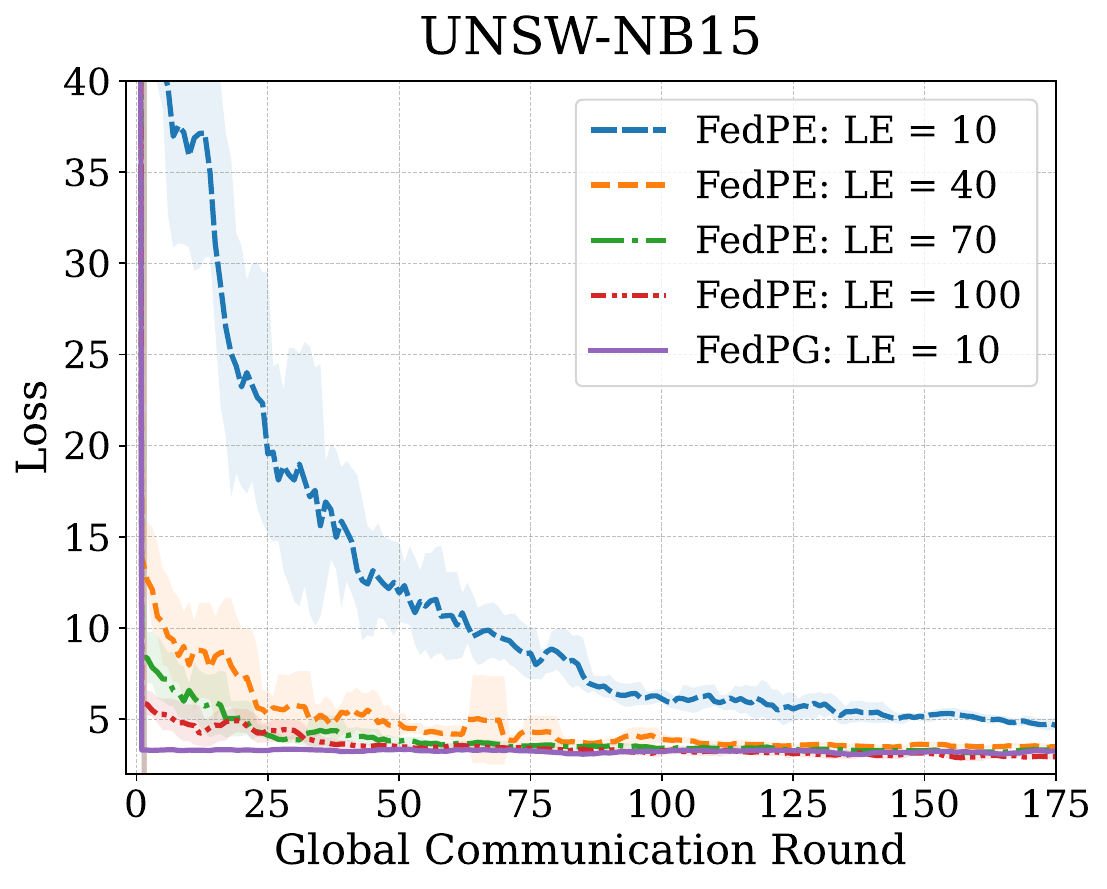}
        \vspace{-15pt}
				\caption{}
		\label{fig:rep2}
	\end{subfigure} 
 	\begin{subfigure}{.24\textwidth}
		\centering
		\includegraphics[width=\textwidth]{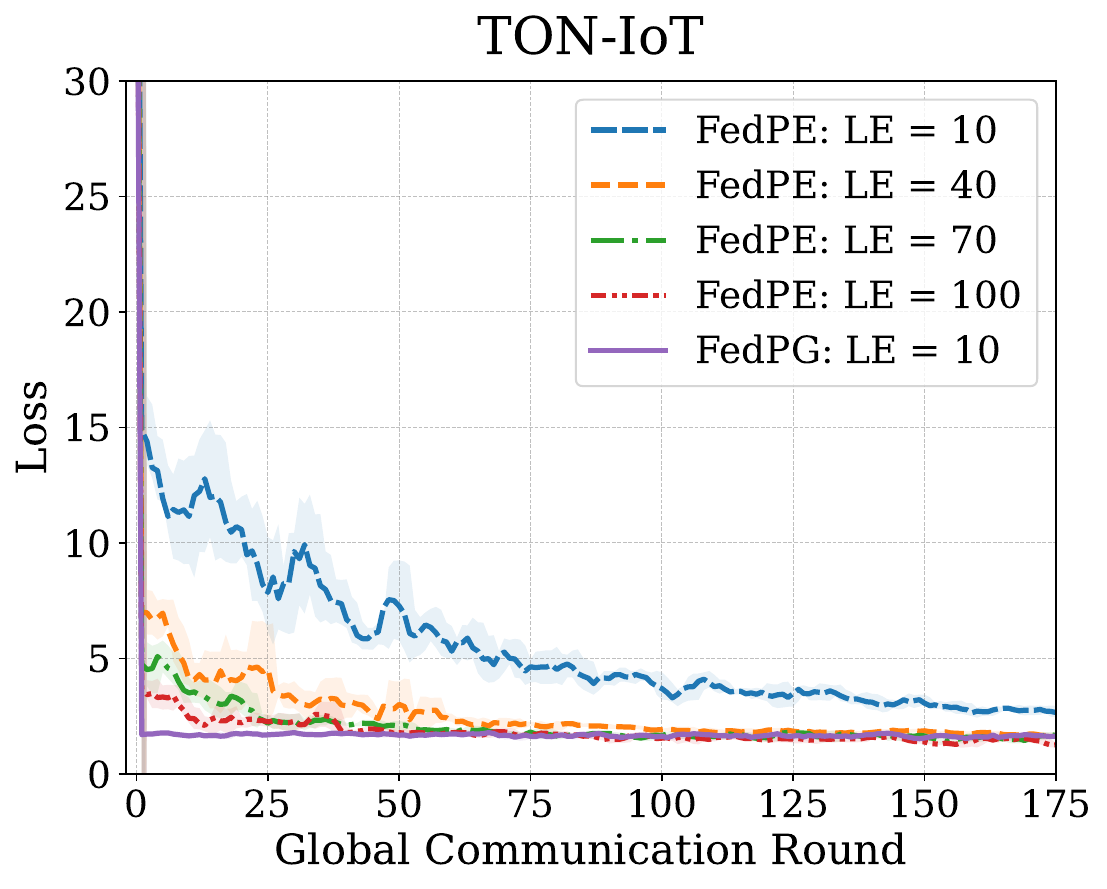}
        \vspace{-15pt}
            \caption{}
		\label{fig:per2}
	\end{subfigure}
	%\label{fig:repochs}
	\caption{(a,b) Effects of $\rho$ on the performance of \textsc{FedPG} and (c,d) Training loss over global iterations}
    \label{fig:finetuning}
	%\vspace{-15pt}
\end{figure*}
\begin{figure*}[h]
	\centering
 	\begin{subfigure}{.24\textwidth}
		\centering
		\includegraphics[width=\textwidth]{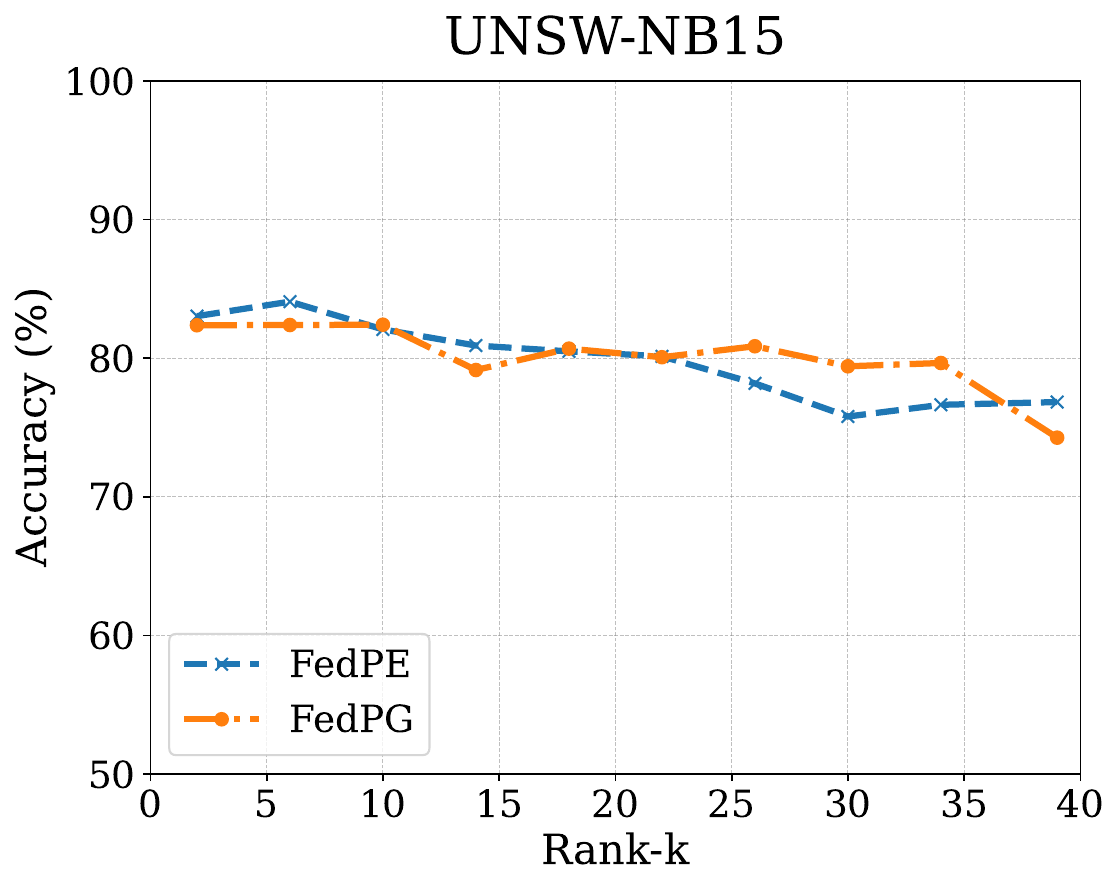}
        \vspace{-15pt}
            \caption{}
		\label{fig:finetuning3}
	\end{subfigure}%
	\begin{subfigure}{.24\textwidth}
		\centering
		\includegraphics[width=\textwidth]{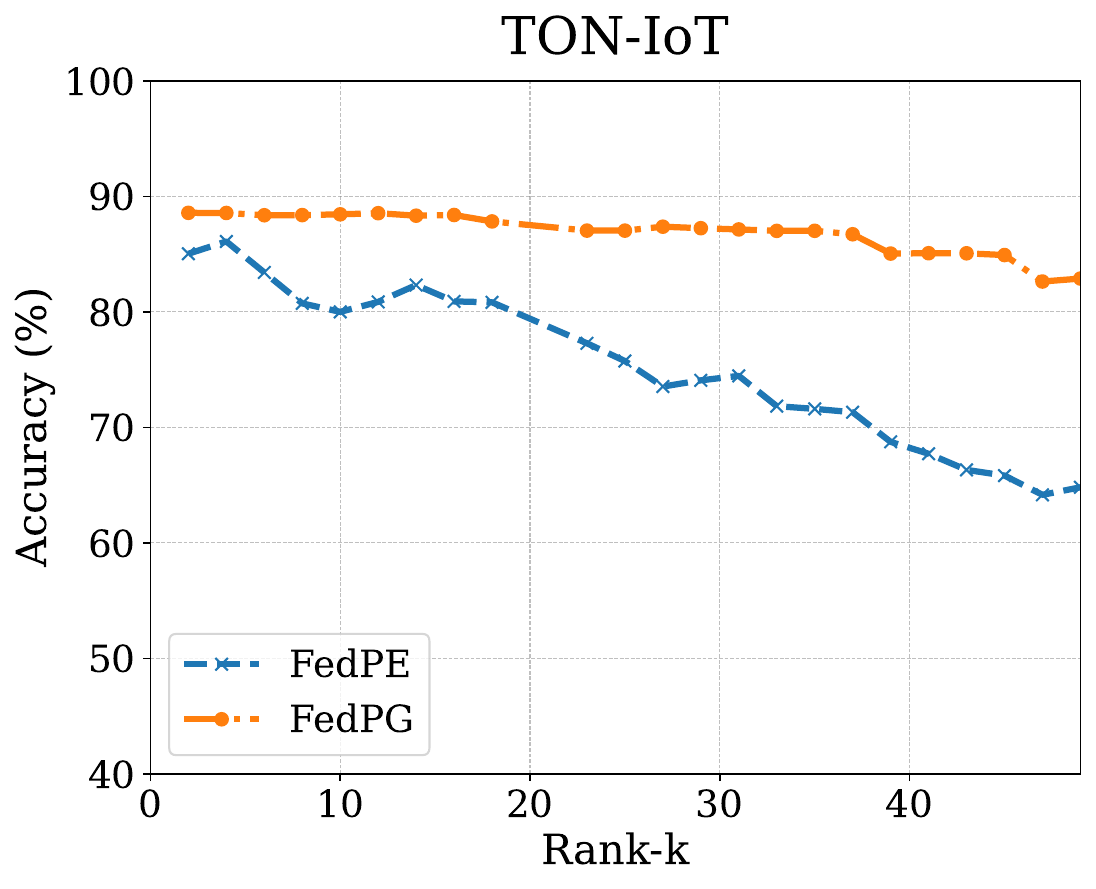}
        \vspace{-15pt}
            \caption{}
		\label{fig:finetuning4}
	\end{subfigure}
 	\begin{subfigure}{.24\textwidth}
		\centering
		\includegraphics[width=\textwidth]{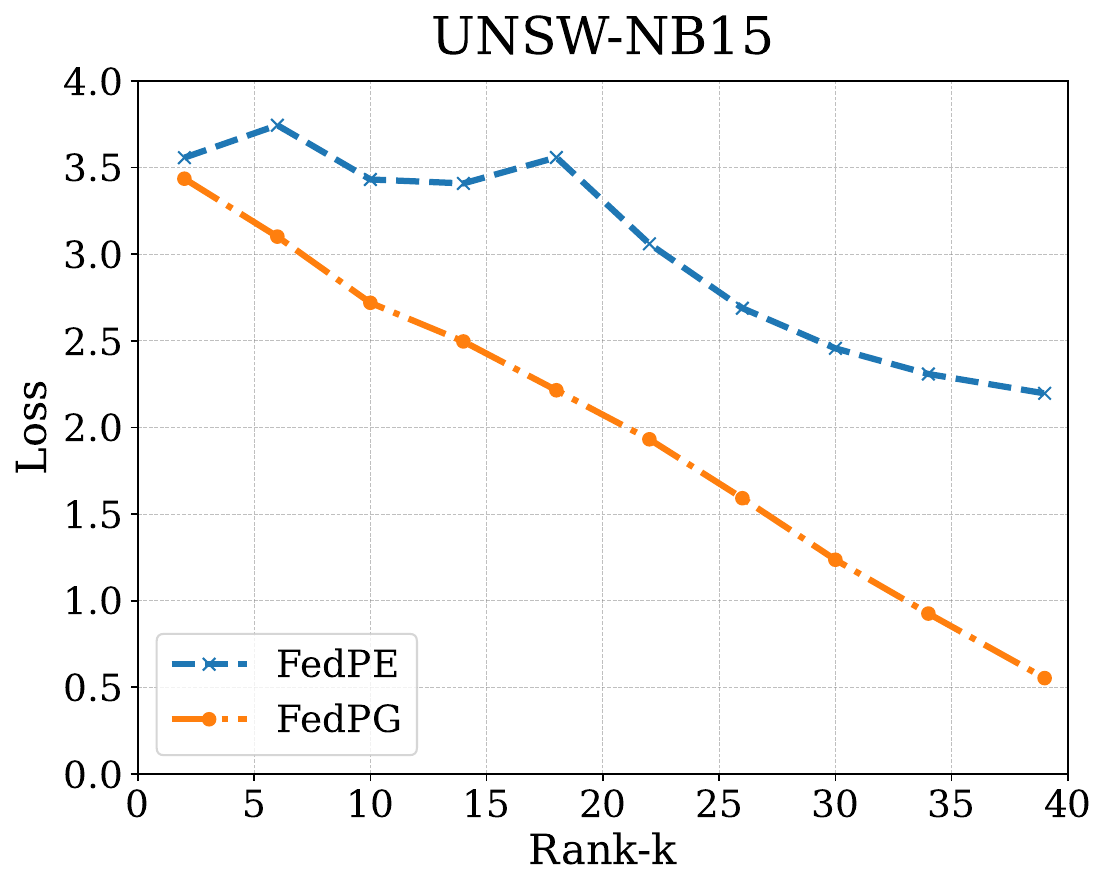}
        \vspace{-15pt}
            \caption{}
		\label{fig:finetuning1}
	\end{subfigure}%
	\begin{subfigure}{.23\textwidth}
		\centering
		\includegraphics[width=\textwidth]{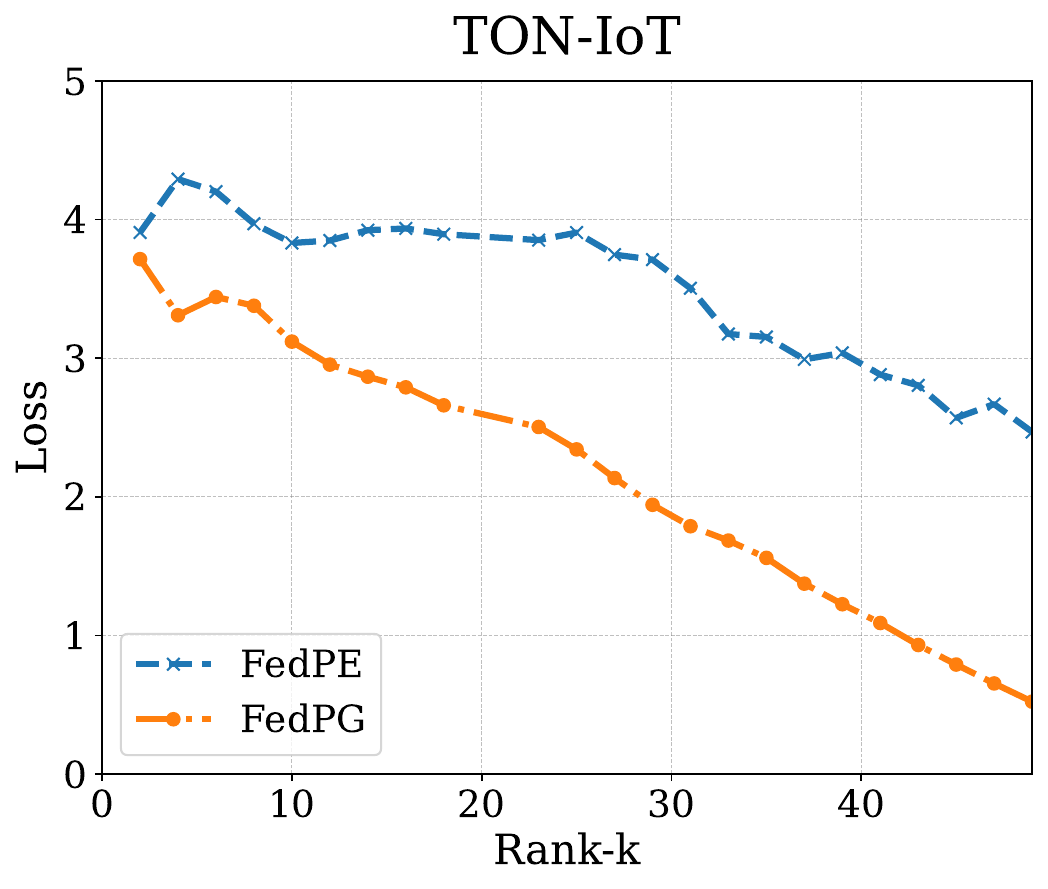}
        \vspace{-15pt}
            \caption{}
		\label{fig:finetuning2}
	\end{subfigure}
	%\label{fig:repochs}
	\caption{Effects of rank-$k$ on the performance of \textsc{FedPE} and \textsc{FedPG}}
    \label{fig:finetuning_rank}
	%\vspace{-15pt}
\end{figure*}
\begin{figure*}[h]
	\centering
	\begin{subfigure}{.24\textwidth}
		\centering
		\includegraphics[width=\textwidth]{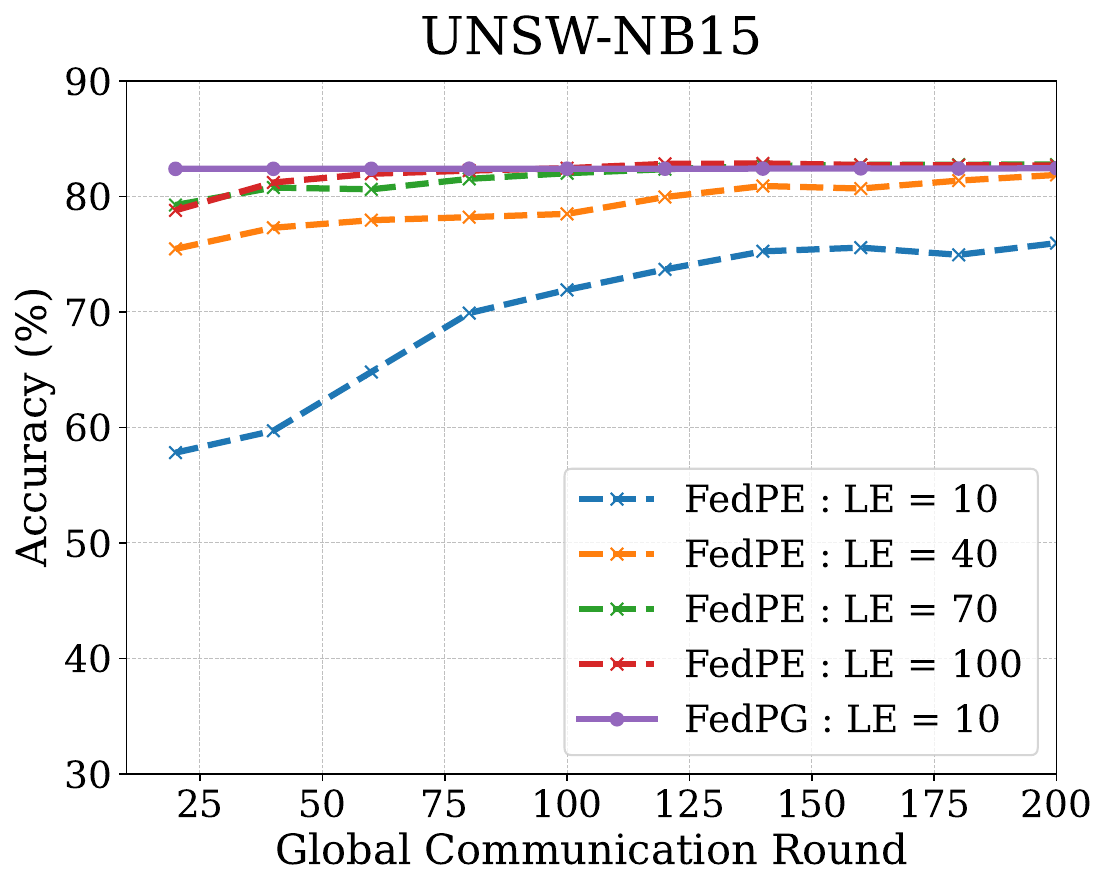}
        \vspace{-15pt}
            \caption{}
		\label{fig:rep3}
	\end{subfigure}%
 	\begin{subfigure}{.23\textwidth}
		\centering
		\includegraphics[width=\textwidth]{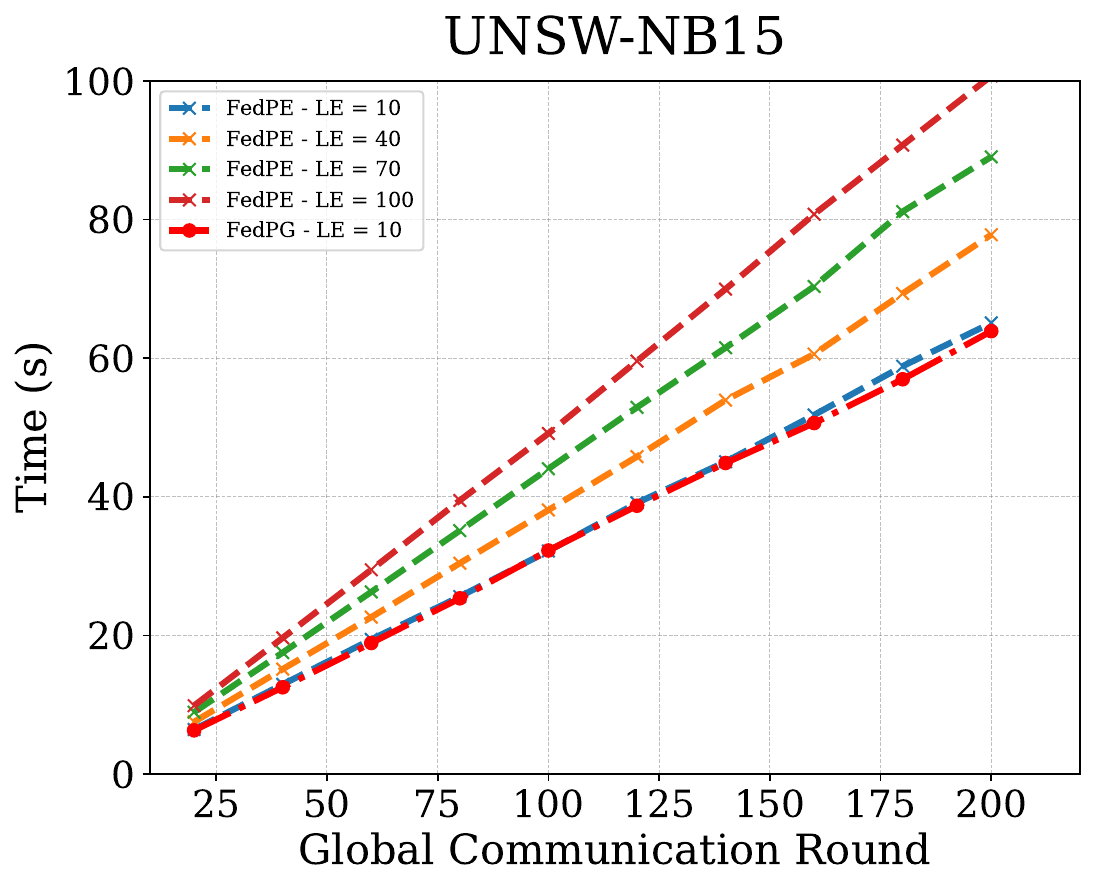}
        \vspace{-15pt}
            \caption{}
		\label{fig:per3}
	\end{subfigure}%
	\begin{subfigure}{.24\textwidth}
		\centering
		\includegraphics[width=\textwidth]{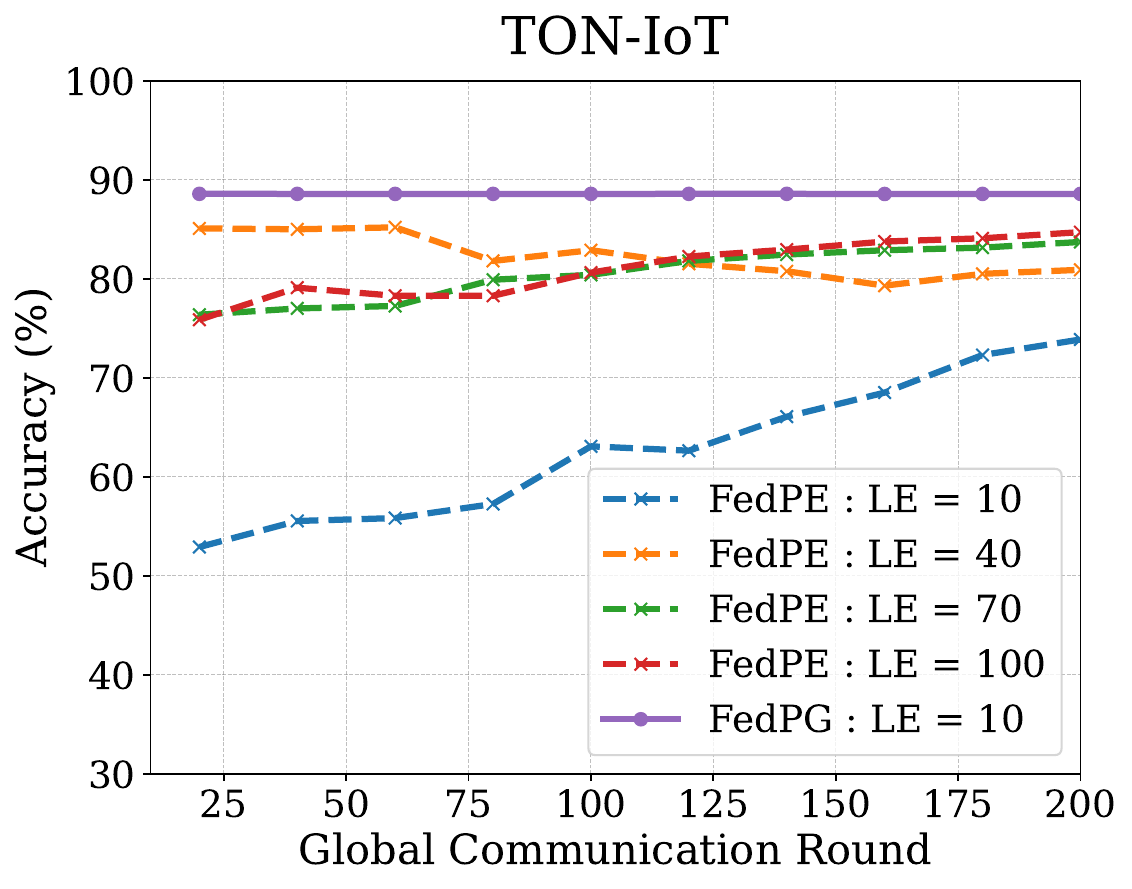}
        \vspace{-15pt}
            \caption{}
		\label{fig:per4}
	\end{subfigure}%
	\begin{subfigure}{.23\textwidth}
		\centering
		\includegraphics[width=\textwidth]{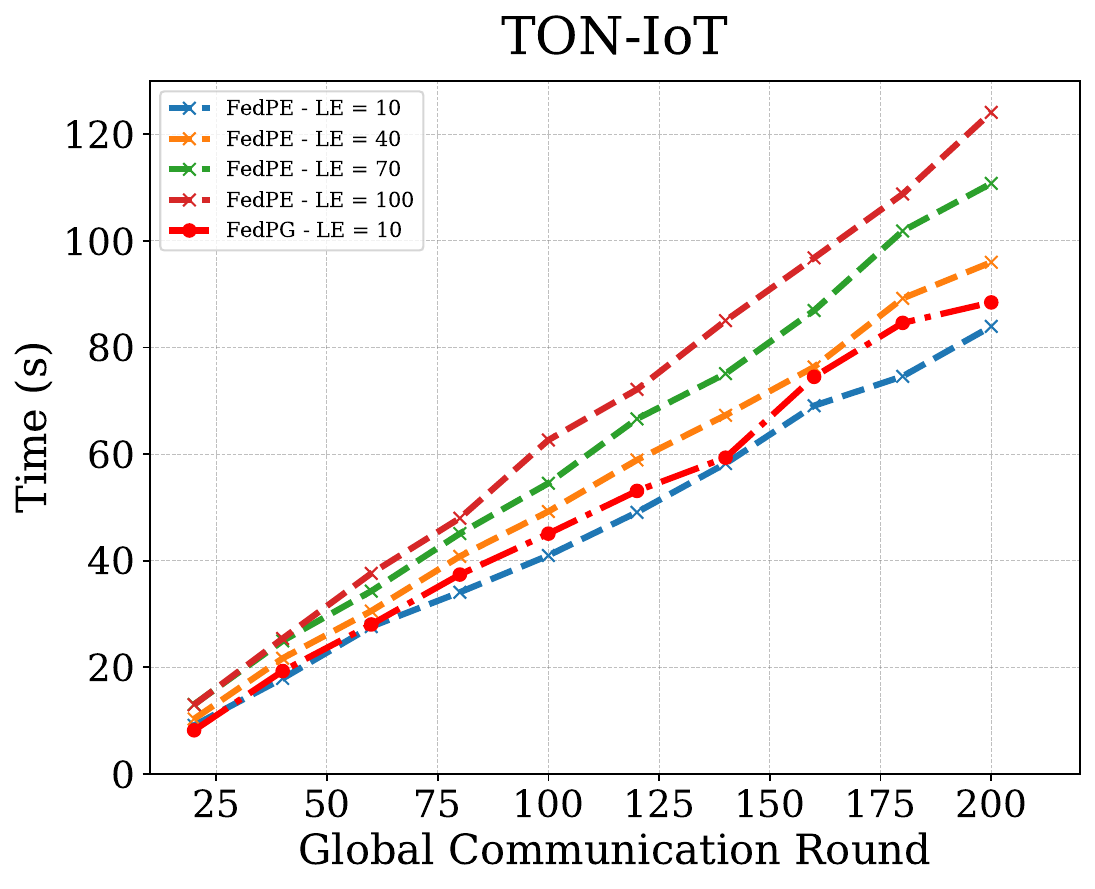}
        \vspace{-15pt}
            \caption{}
		\label{fig:per5}
	\end{subfigure}
	%\label{fig:repochs}
	\caption{Test accuracy and training time over global iterations of \textsc{FedPE} and \textsc{FedPG} on UNSW-NB15 (a,b) and TON-IoT (c,d)}
    \label{fig:convergence}
	%\vspace{-15pt}
\end{figure*}

In our analysis, we scrutinize the convergence efficacy of our Federated PCA algorithms — \textsc{FedPE} (Federated PCA over Euclidean Space) and \textsc{FedPG} (Federated PCA over Grassmann Manifold) — as depicted in Fig.\ref{fig:finetuning}, Fig.~\ref{fig:convergence} and Fig.~\ref{fig:finetuning_rank}.  Notably, for the same number of local communication rounds ($LE=10$), FedPG displays a significantly faster convergence compared to FedPE. It's worth noting that \textsc{FedPE} can achieve quicker convergence by augmenting the value of $LE$, although this also extends the training duration. These findings underscore \textsc{FedPG}'s ability to deliver robust detection performance with decreased training time, an attribute highly sought after in practical IDS development contexts. Fig.\ref{fig:finetuning5}, \ref{fig:finetuning6} illustrate the precise performance of \textsc{FedPG} across various thresholds. In our study, we uniformly set $\rho=1.0$ across all methods and investigate their detection performance by various metrics.

%\blue{
The impact of low-rank approximation is presented in Fig.~\ref{fig:finetuning3} and ~\ref{fig:finetuning4}. Surprisingly, increasing the rank-k does not necessarily lead to an improvement in the performance of both \textsc{FedPG} and \textsc{FedPE}, as observed in both datasets. This suggests that our PCA-based anomaly detection algorithms can maintain a low memory footprint and operate efficiently with low computational complexity, making them ideal for resource-constrained environments. Despite \textsc{FedPG} and \textsc{FedPE} providing better approximations for a given dataset with larger rank-k, they also tend to precisely recover data points lying in high-dimensional space. However, high-dimensional data samples often act as outliers, resembling attacks at times. This creates a dilemma between achieving optimal performance and accurate approximation, as seen in the loss-accuracy contrast presented in Fig.~\ref{fig:finetuning1}, \ref{fig:finetuning2}, and Fig.~\ref{fig:finetuning3}, \ref{fig:finetuning4}. Abnormal detection using PCA relies on measuring the distance between data points and principal components, making it challenging to differentiate between high-dimensional data points and malicious data.
%}

To examine the effectiveness of \textsc{FedPG} against \textsc{FedPE} in terms of computational complexity, we investigate the training time of both algorithms on the same machine. While Fig. \ref{fig:rep3} and \ref{fig:per4} demonstrate that \textsc{FedPE} surpasses \textsc{FedPG} in terms of the total number of communication rounds, our investigation reveals that \textsc{FedPG} is more effective in managing computational costs, as depicted in Fig. \ref{fig:per3} and \ref{fig:per5}. It is worth noting that \textsc{FedPG} sacrifices a small amount of computational resources in exchange for significantly higher performance.

\subsubsection{Effects of number clients on performance}

\begin{figure}[t]
	\centering
	\begin{subfigure}{.25\textwidth}
		\centering
		\includegraphics[width=\textwidth]{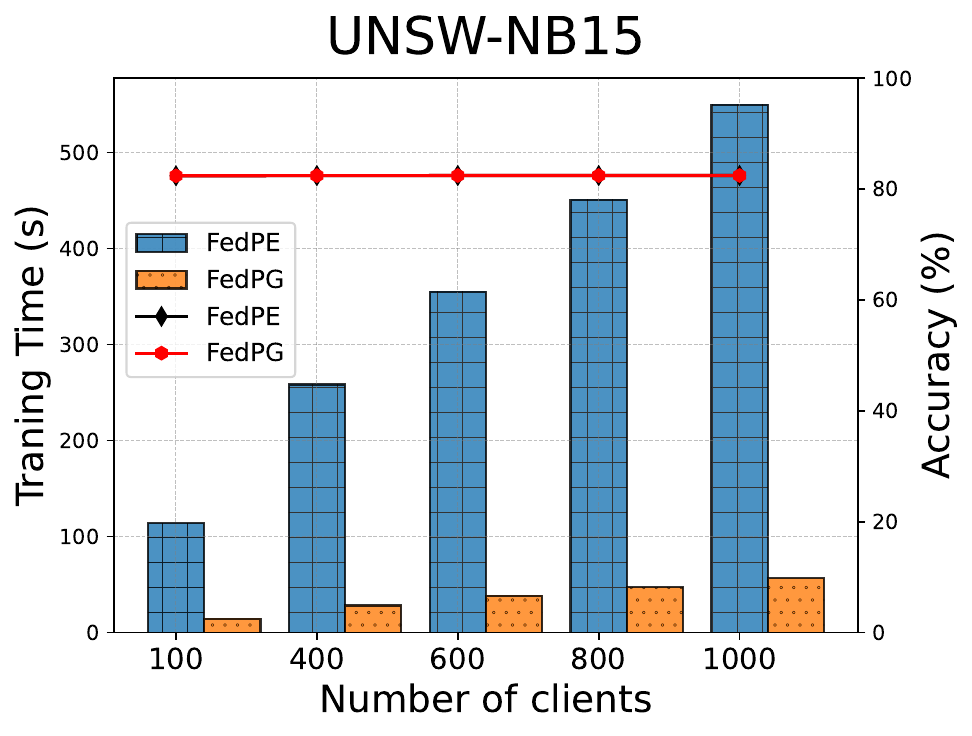}
        %\vspace{-15pt}
            \caption{}
		\label{fig:per1}
	\end{subfigure}%
	\begin{subfigure}{.25\textwidth}
		\centering
		\includegraphics[width=\textwidth]{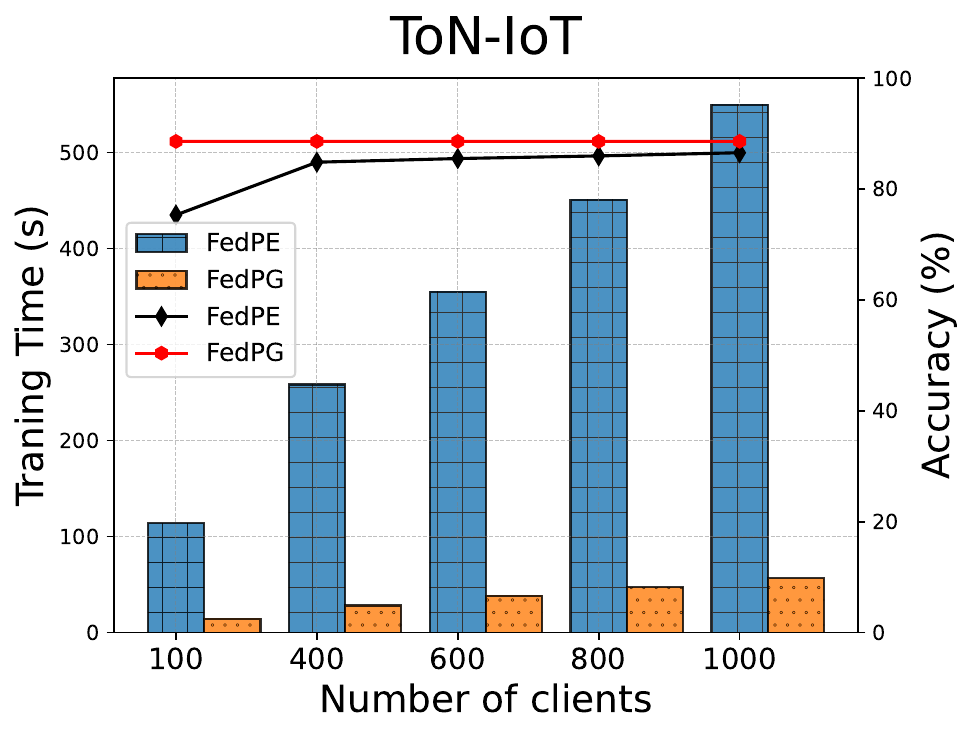}
        %\vspace{-15pt}
            \caption{}
		\label{fig:per2}
	\end{subfigure}
	\caption{Effects of number of clients on training time and performance of \textsc{FedPE} and \textsc{FedPG}}
    \label{fig:no_clients}
	%\vspace{-15pt}
\end{figure}

We evaluate the performance of \textsc{FedPE} and \textsc{FedPG} on different scales of IoT systems by varying the number of clients. The results illustrated in Fig. \ref{fig:no_clients} reveal a noticeable trade-off between training time and accuracy for these algorithms.

For both datasets, \textsc{FedPG} consistently demonstrated a significantly shorter training time as compared to \textsc{FedPE}, regardless of the number of clients. For example, on the TON-IoT dataset, the training time of \textsc{FedPG} was less than a sixth of the training time of \textsc{FedPE} for the same number of clients. In contrast, the accuracy of \textsc{FedPG} remained relatively constant and independent of the number of clients in both datasets. \textsc{FedPE}, on the other hand, demonstrated an improvement in accuracy as the number of clients increased. However, even with this improvement, it was unable to reach the accuracy of \textsc{FedPG} on the TON-IoT dataset. For the UNSW-NB15 dataset, both algorithms performed comparably in terms of accuracy, with \textsc{FedPE} even slightly surpassing \textsc{FedPG} in fewer clients, but then slightly underperforming in more clients.

These results suggest that \textsc{FedPG} could offer substantial time savings over \textsc{FedPE}  without compromising detection accuracy, particularly for large numbers of clients. However, the appropriate choice of algorithm may also depend on the specific requirements and constraints of the task and system at hand, such as the importance of speed versus detection accuracy and the number of clients in the system.

\begin{figure*}[t]
	\centering
	\begin{subfigure}{.24\textwidth}
		\centering
		\includegraphics[width=\textwidth]{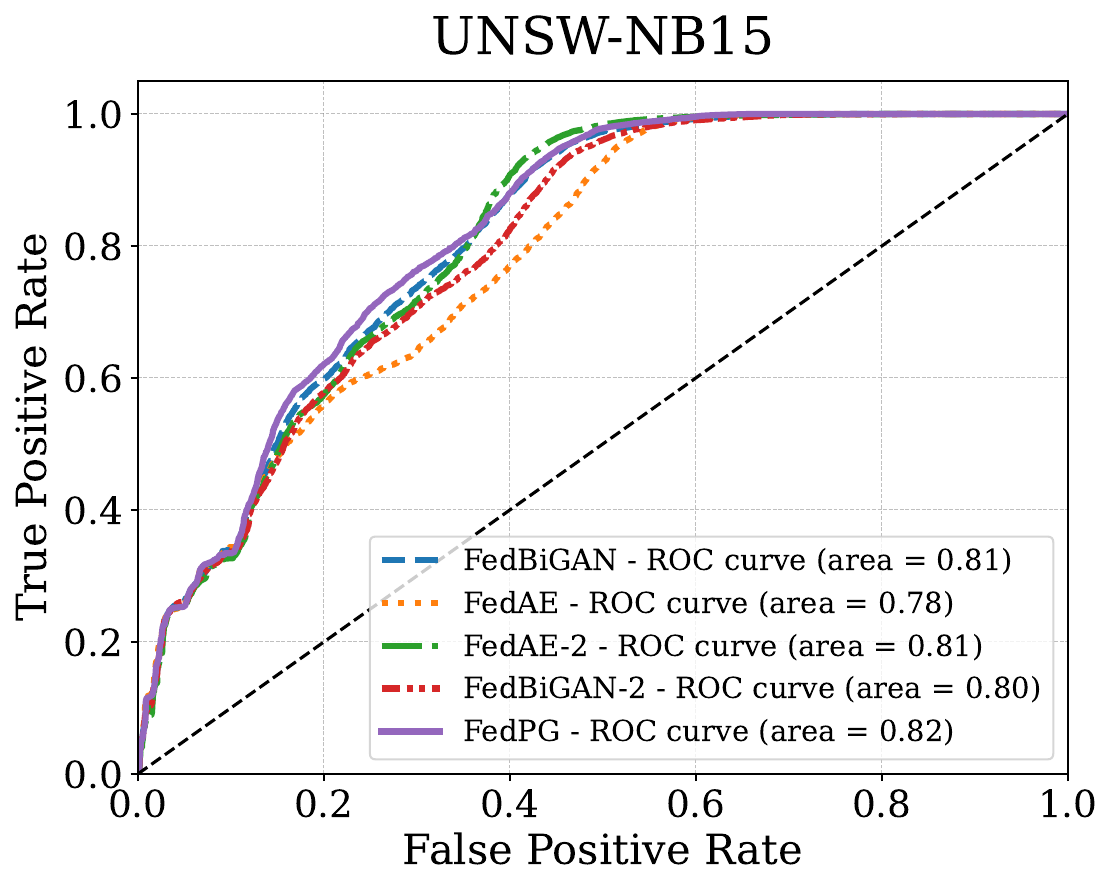}
        %\vspace{-15pt}
            \caption{}
		\label{fig:roc1}
	\end{subfigure}%
 	\begin{subfigure}{.24\textwidth}
		\centering
		\includegraphics[width=\textwidth]{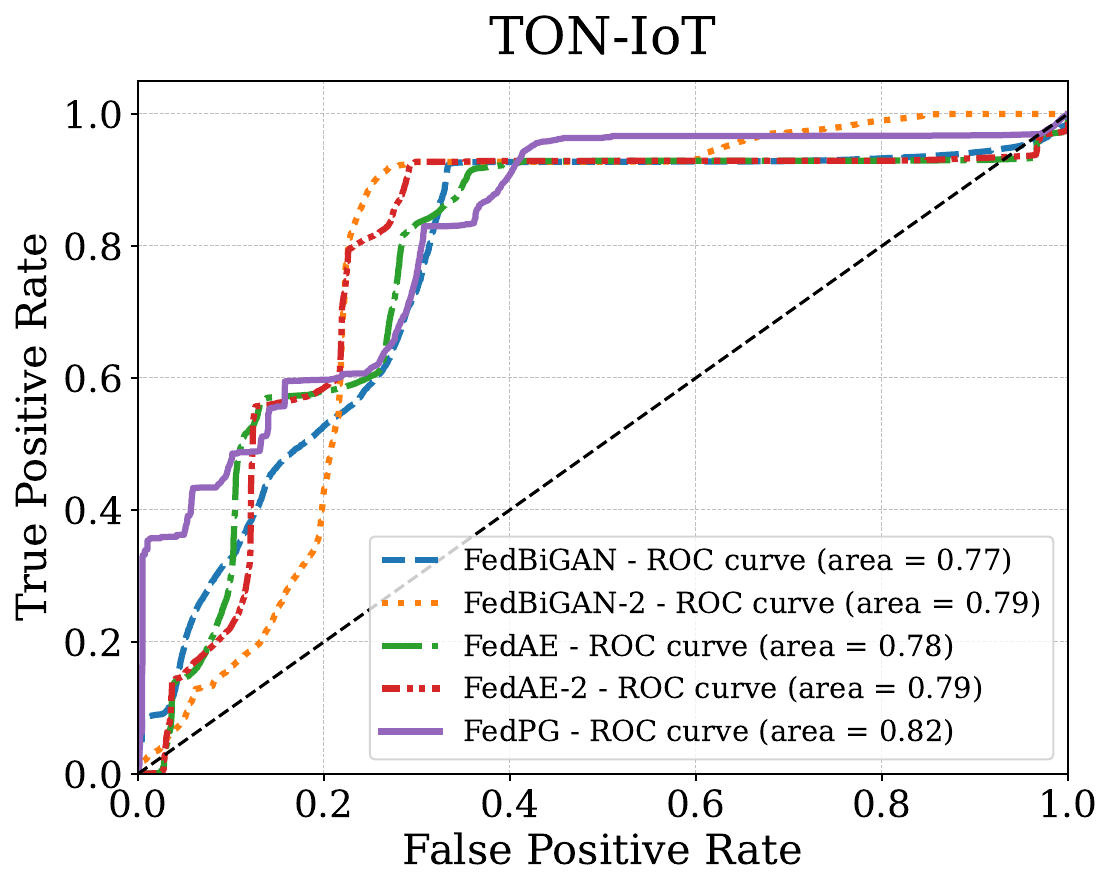}
        %\vspace{-15pt}
            \caption{}
		\label{fig:roc2}
	\end{subfigure}
 	\begin{subfigure}{.24\textwidth}
		\centering
		\includegraphics[width=\textwidth]{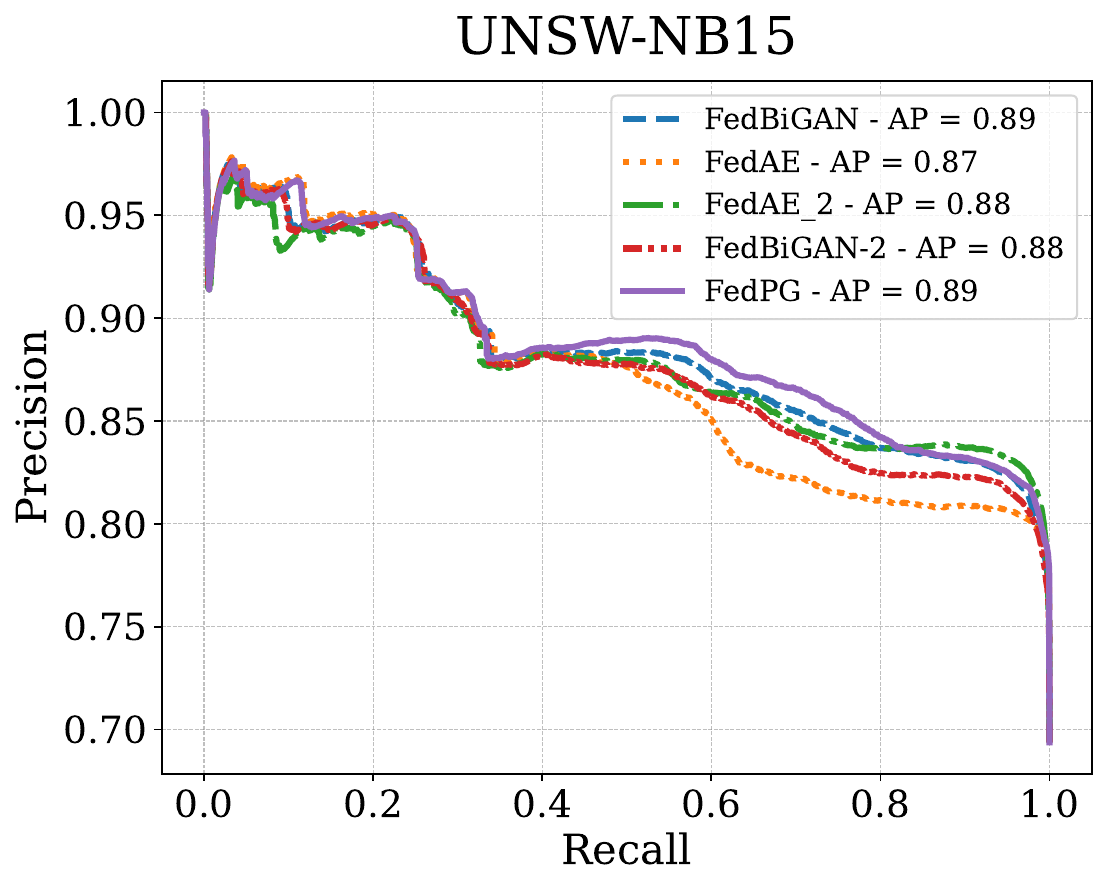}
        %\vspace{-15pt}
            \caption{}
		\label{fig:pre1}
	\end{subfigure}%
 	\begin{subfigure}{.24\textwidth}
		\centering
		\includegraphics[width=\textwidth]{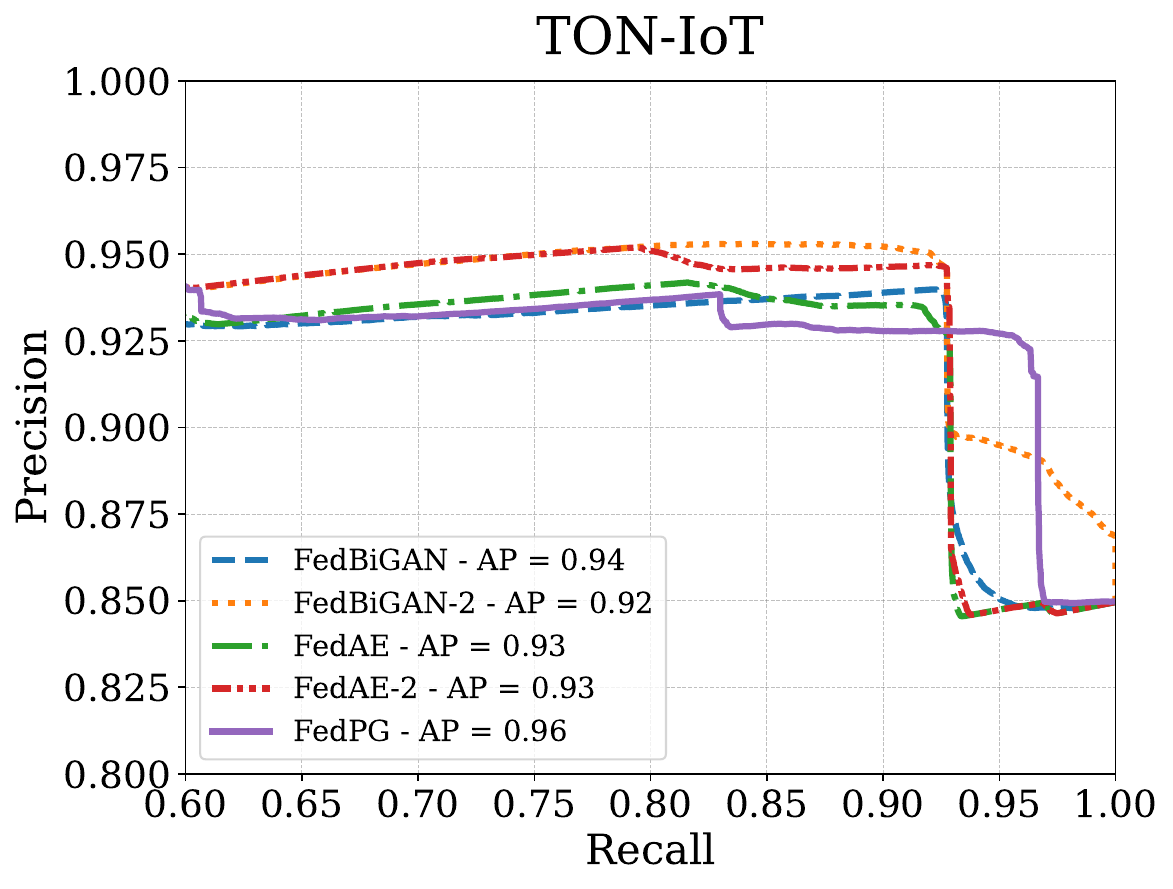}
        %\vspace{-15pt}
            \caption{}
		\label{fig:pre2}
	\end{subfigure}
	\caption{(a,b) Receiver Operating Characteristic (ROC) curve and (c,d) Precision-Recall curve}
  	\label{fig:roc_pre}
	%\vspace{-15pt}
\end{figure*}

\begin{figure*}[h]
	\centering
	\begin{subfigure}{.24\textwidth}
		\centering
		\includegraphics[width=\textwidth]{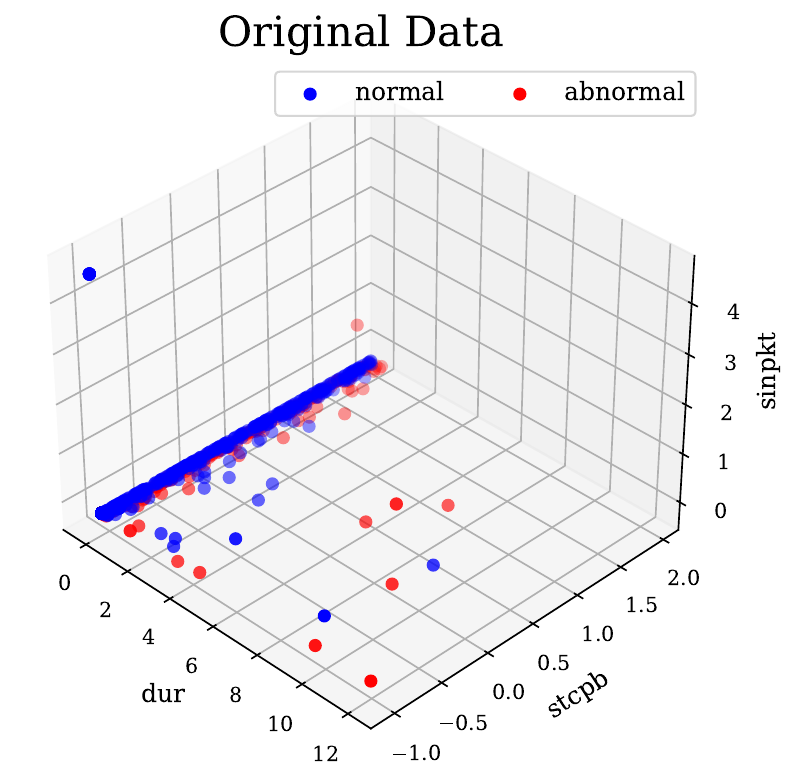}
        %\vspace{-15pt}
            \caption{}
		\label{fig:rep2}
	\end{subfigure} 
	\begin{subfigure}{.24\textwidth}
		\centering
		\includegraphics[width=\textwidth]{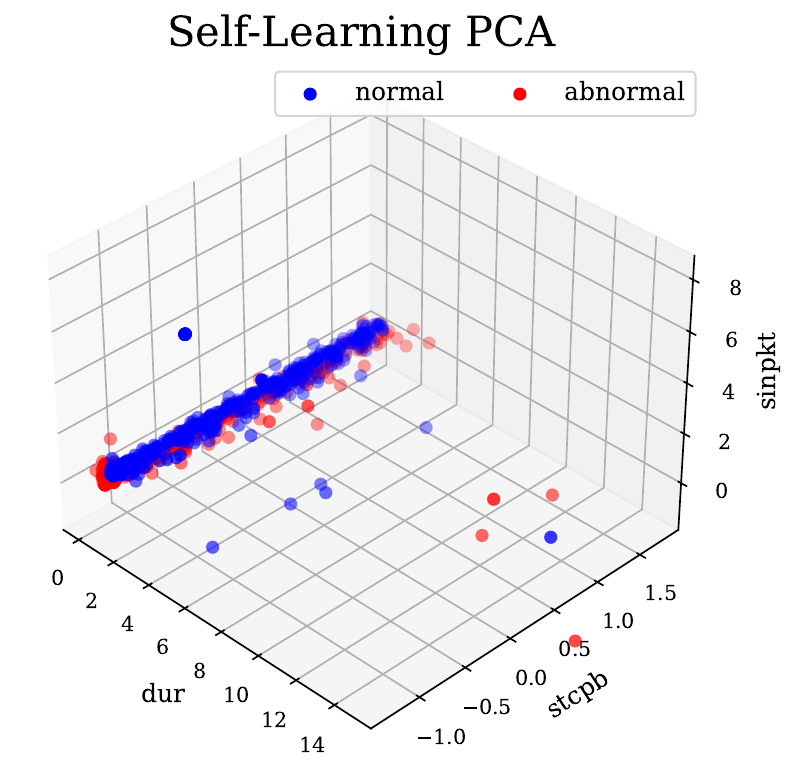}
        %\vspace{-15pt}
            \caption{}
		\label{fig:rep1}
	\end{subfigure}%
 	\begin{subfigure}{.24\textwidth}
		\centering
		\includegraphics[width=\textwidth]{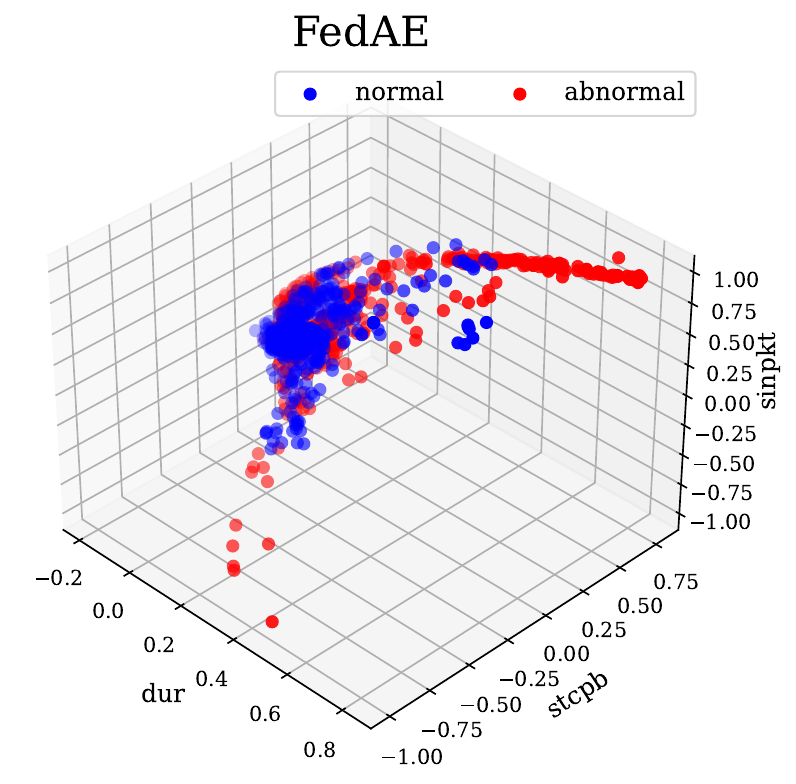}
        %\vspace{-15pt}
            \caption{}
		\label{fig:rep1}
	\end{subfigure}%
	\begin{subfigure}{.24\textwidth}
		\centering
		\includegraphics[width=\textwidth]{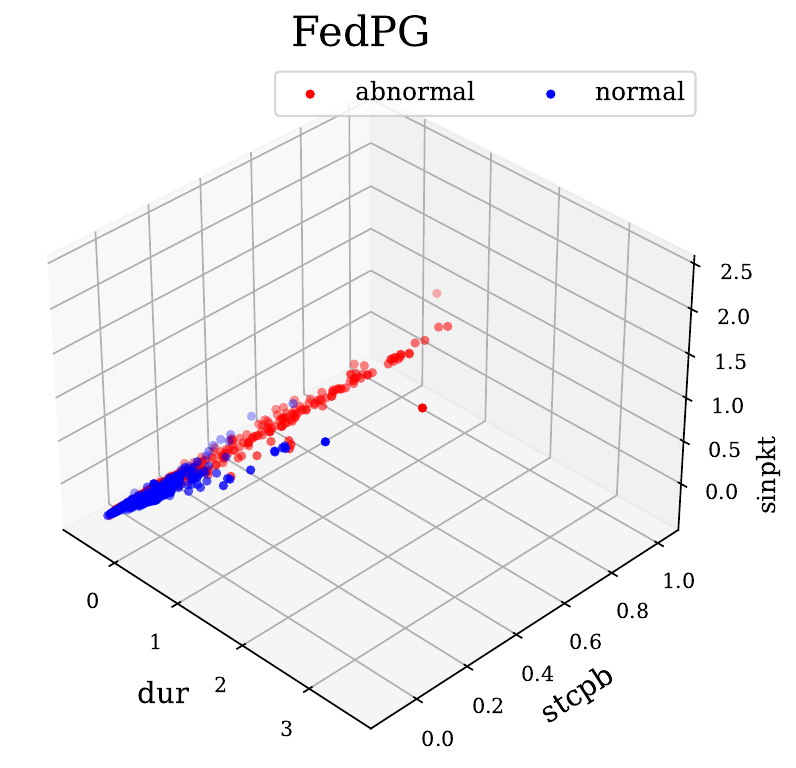}
        %\vspace{-15pt}
            \caption{}
		\label{fig:rep2}
	\end{subfigure}
	%\label{fig:repochs}
	\caption{Visualization of UNSW-NB15 traffic after reconstruction using different methods}
    \label{fig:visualization}
	%\vspace{-15pt}
\end{figure*}

\begin{table*}[h]
	\centering
	\caption{Performance of network anomaly detection tasks on UNSW-NB15 and TON-IoT datasets.}
	\label{tab:performance}
    \renewcommand{\arraystretch}{1.15} % For row height
    \scalebox{0.9}{
	\begin{tabular}{|c||c|c|c|c|c||c|c|c|c|c|}
		\hline
        \textbf{Dataset} & \multicolumn{5}{c||}{\textbf{UNSW-NB15}} & \multicolumn{5}{c|}{\textbf{TON-IoT}} \\
        \hline
		\textbf{Method}            & \textbf{Acc} & \textbf{Pre} & \textbf{Recall} & \textbf{F1-Score} & \textbf{FNR} & \textbf{Acc} & \textbf{Pre} & \textbf{Recall} & \textbf{F1-Score} & \textbf{FNR} \\ 
		\hline
        \hline
		%\midrule
  		Self-learning PCA & 63.80  & 83.19 & 59.94  & 69.68 &  40.06 & 51.81  & 86.79 & 51.07  & 64.30 &  48.93  \\
        %\hline
		%Centralized PCA   & --  & -- & --  & -- &  -- & --  & -- & --  & -- &  --  \\ 
		%\midrule
		%\midrule
		%Centralized AE      & 80.97   & 90.93  & 80.61 & 85.47 & 19.39
        %                    & 87.94  & 94.71  & 90.88  & 92.76 & 9.12 \\
  		%Centralized BiGAN   & 80.28   & 85.17  & 86.67  & 85.91 & 13.33  
        %                    & \textbf{89.04}  & 94.24  & 92.77  & 93.50 & 7.22  \\
		%\midrule
        \hline
		FedAE      & 80.88   & 80.66  & \underline{95.27}  & 87.36 & \underline{4.73} & 
                     87.28   & 93.52  & 91.36  & 92.43 & 8.64  \\
		FedAE-2     & \underline{82.80}   & \underline{83.45}  & 93.80  & \underline{88.32} & 6.20 & 
                     \underline{89.09}   & 94.68  & 92.34  & 93.50 & 7.66  \\
        FedBiGAN    & 81.21  & 83.04  & 91.63  & 87.12 & 8.36
                    & 88.53  & 93.98  & 92.42  & 93.19 & 7.57  \\
        FedBiGAN-2  & 81.59  & 81.98  & 94.16  & 87.65 & 5.84
                    & 89.08  & \underline{95.05}  & 91.93  & 93.47 & 8.07  \\
  %\midrule
		%Fed-AE           & 80.7   & 87.62  & 76.96  & 14.36 &  81.94  \\
		%\hline
		%Fed-LSTM         & 75.53  & 90.16  &   & 9.24 &  74.87  \\
		%\hline
  		%\midrule
        \hline
		{\textbf{\textsc{FedPE} (ours)}}   & {\textbf{82.15}} & {\textbf{81.06}} & {\textbf{95.14}} & {\textbf{87.53}}     & {\textbf{6.25}} & {\textbf{87.06}} & {\textbf{89.82}} & {\underline{\textbf{94.94}}} & {\textbf{92.31}} & {\textbf{6.13}} \\        
		\textbf{\textsc{FedPG} (ours)}    & \textbf{81.95} & \textbf{82.82} & \textbf{93.36} & \textbf{87.77}     & \textbf{6.63} & \textbf{88.94} & \textbf{92.79} & \textbf{94.32} & \underline{\textbf{93.55}}     & \underline{\textbf{5.68}} \\ 
        \hline
	\end{tabular}}
\end{table*}

\subsubsection{Performance of Network Anomaly Detection Tasks}
%The results presented \ref{tab:performance} demonstrate the superior performance of \textsc{FedPG} over the existing methods for network anomaly detection tasks on both UNSW-NB15 and ToN-IoT datasets. When compared to Self-learning PCA, FedAE, and FedBiGAN, the \textsc{FedPG} algorithm consistently achieved higher accuracy, precision, and F1-Score, while maintaining a lower false negative rate. Particularly noteworthy is the improvement on the ToN-IoT dataset, where FedPG attained a precision of 92.79\%, an impressive recall of 94.32\%, and a remarkable F1-Score of 93.55\%, outperforming all other baselines. The lower false negative rate of FedPG, especially 5.68\% on the ToN-IoT dataset, as compared to the highest among other methods, which is 8.94\% by FedAE, further underscores its effectiveness in accurately identifying network anomalies. These results demonstrate the robustness and reliability of FedPG as a solution for network anomaly detection, and suggest its potential for wider application in cybersecurity tasks.

We evaluate the performance of \textsc{FedPG} alongside other baselines in the context of network anomaly detection on both the UNSW-NB15 and TON-IoT datasets. For all methods, we establish optimal thresholds to delineate normal from anomalous behavior using the AUC-ROC curve corresponding to their reconstruction errors. This approach aims to strike a balance by maximizing TPR and minimizing FPR, ensuring robustness and precision in anomaly detection. Such an evidence-driven thresholding strategy underscores the reliability of the reconstruction error as a metric for detecting anomalies in network traffic.  

%\blue{
As indicated in Table~\ref{tab:performance}, Self-learning PCA, while offering a foundational unsupervised technique, lags behind in performance when compared to other methods. This outcome aligns with expectations, given the diverse and highly heterogeneous nature of clients' data. The performance of \textsc{FedPG} and \textsc{FedPE} is comparable and slightly better than FedAE and FedBiGAN. Additionally, they demonstrate a competitive performance against more sophisticated non-linear structures such as FedAE-2 and FedBiGAN-2. It should be noted that while \textsc{FedPG} may not outperform all other techniques across all metrics, it does exhibit a commendable balance of measures.
%}

On the UNSW-NB15 dataset, \textsc{FedPG} not only shows a high accuracy (81.95\%) but also balances this with a relatively low false negative rate (6.63\%), indicating a robust ability to correctly classify normal behavior while minimizing the chance of missing true anomalies. Similarly, on TON-IoT, \textsc{FedPG} holds its own by showcasing the highest recall (94.32\%), reflecting its superior ability to correctly identify anomalous events. Moreover, it has the lowest false negative rate (5.68\%)  among the evaluated methods, which underscores its performance stability. Furthermore, the linear nature of \textsc{FedPG} simplifies model interpretability, often leading to more computationally and communication-efficient solutions while still being able to capture significant patterns in the data for network anomaly detection. Meanwhile, although FedAE, FedAE-2, FedBiGAN, and FedBiGAN-2 can potentially capture more complex relationships in the data, they tend to require more computational and memory resources, harder to interpret and may not always result in better performance. This suggests that the balanced performance of our linear method \textsc{FedPG} is an attractive alternative for real-world IDS where simplicity, efficiency, and interpretability are crucial.

The superior performance of \textsc{FedPG} is further validated by the area under the Receiver Operating Characteristic (AUC-ROC) and Precision-Recall curves, depicted in Fig. \ref{fig:roc_pre}. On both the UNSW-NB15 and TON-IoT datasets, \textsc{FedPG} demonstrates higher AUC-ROC values of 0.82, surpassing other baselines (Fig. \ref{fig:roc1}, \ref{fig:roc2}), indicating its enhanced ability to maintain high true positive rates while reducing false positive rates. This performance superiority is corroborated by the Precision-Recall curves, with \textsc{FedPG} achieving an Average Precision (AP) of 0.89 and 0.96 on UNSW-NB15 and TON-IoT datasets respectively (Fig. \ref{fig:pre1}, \ref{fig:pre2}), comparable with the competing methods. These results collectively affirm the robustness of \textsc{FedPG} in achieving high precision at various levels of recall, underscoring its potential as an effective solution for network anomaly detection.

To delve deeper into the efficacy of the suggested technique, we employed 3D scatter plots to illustrate the transformation of UNSW-NB15 traffic logs as per the techniques described, as depicted in Fig.\ref{fig:visualization}. For the purpose of plotting, we chose three features\footnote{Network features in Fig.\ref{fig:visualization}: duration ('dur'), source TCP base sequence number ('stcpb'), source interpacket arrival time ('sinpkt')}, following which we reconstructed the original logs from the latent representation discerned by Self-Learning PCA, FedAE, and \textsc{FedPG}. It's evident that \textsc{FedPG} essentially repositions original logs from an established coordinate framework into an alternate one, proficient in distinguishing between regular logs and anomalies. This affirms that when \textsc{FedPG} redevelops the original logs from the primary components, it results in a minimal reconstruction discrepancy for regular data but a substantial error.

\subsubsection{Communication Efficiency}

\begin{figure}[t]
	\centering
 	%\begin{subfigure}{.6\linewidth}
		\centering
		\includegraphics[width=0.6\columnwidth]{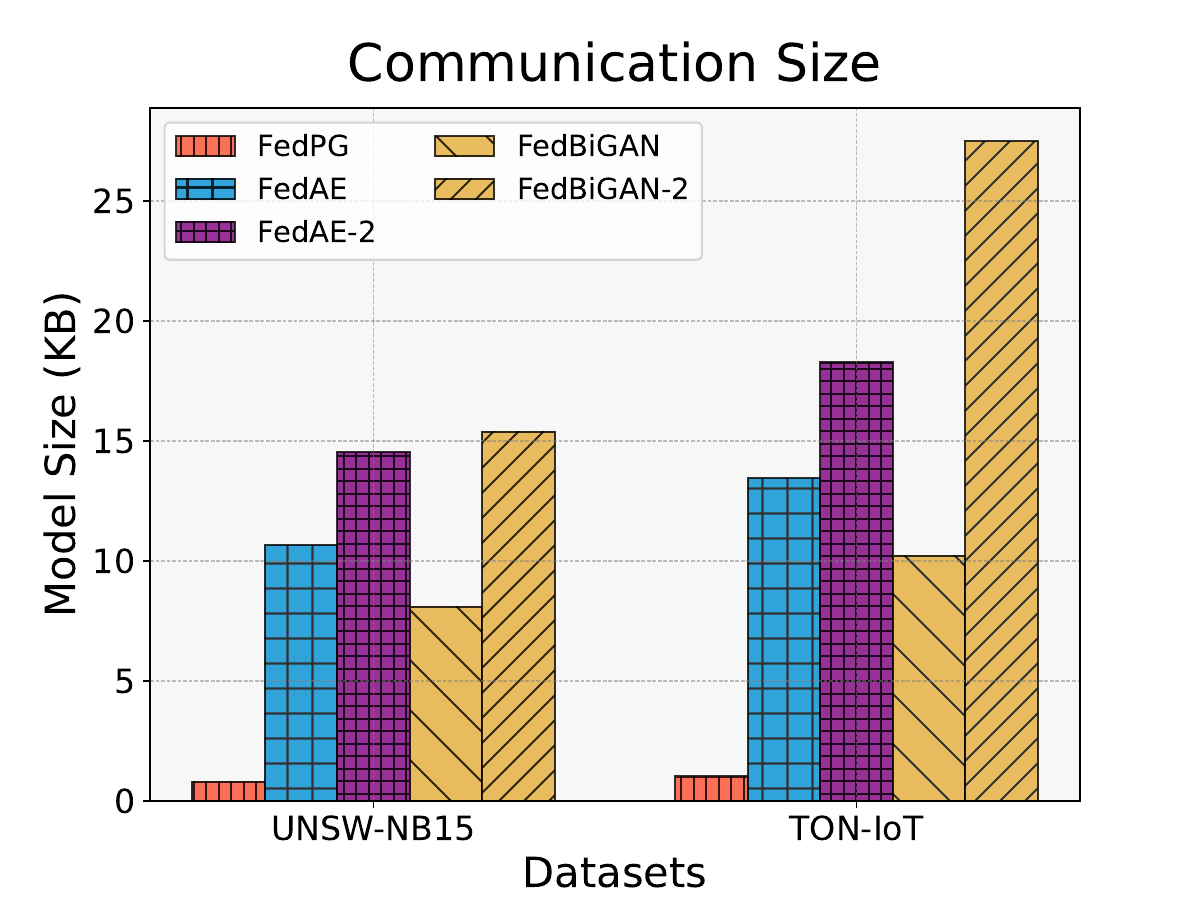}
        %\vspace{-15pt}
		\caption{The size of model exchanged in each communication round.}
		\label{fig:size}
	%\end{subfigure}%
\end{figure}
\begin{figure}[t]
	%\begin{subfigure}{.6\linewidth}
		\centering
		\includegraphics[width=0.55\columnwidth]{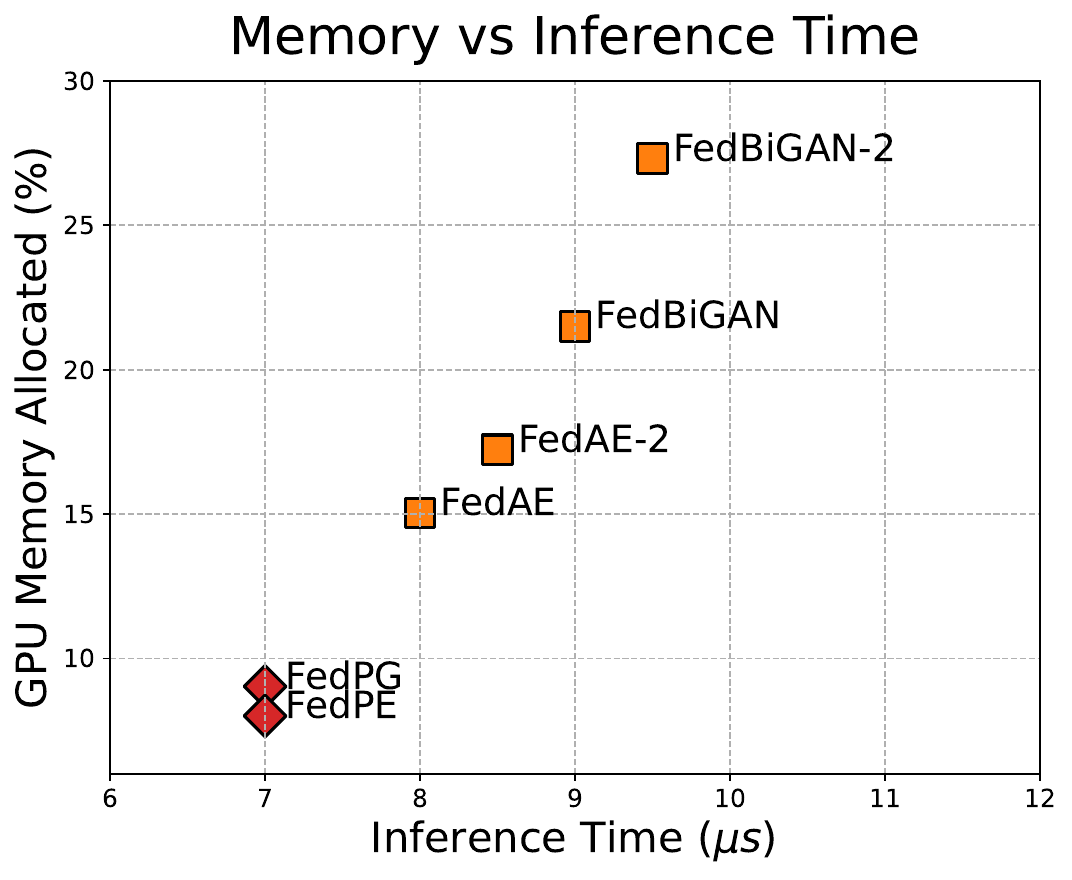}
        %\vspace{-15pt}
		%\captionof{figure}{}
        \caption{The memory consumption and inference time of all methods.}
		\label{fig:memory}
	%\end{subfigure}
     
     %\label{fig:size_memory}
	%\vspace{-15pt}
\end{figure}
In the realm of IoT anomaly detection, the transmission size of models during communication rounds plays a pivotal role, directly influencing bandwidth and overall efficiency. Our study reveals a significant communication advantage of \textsc{FedPG} when juxtaposed against FedAE, FedBiGAN, FedAE-2, and FedBiGAN-2. Specifically, when considering the UNSW-NB15 datasets, \textsc{FedPG} boasts a communication efficiency that's 6.7 and 10 times superior to FedAE and FedBiGAN respectively. While FedAE-2 and FedBiGAN-2 may have a slight edge performance-wise, their demand for bandwidth is significantly higher, necessitating model sizes 10 and 15 times that of \textsc{FedPG}.
%We evaluate the size of the model that needs to be transmitted at each communication round during the training phase for each method. Fig. \ref{fig:size} starkly underscores the pronounced communication efficiency of \textsc{FedPG} in comparison to FedAE, FedBiGAN, FedAE-2 and FedBiGAN-2. In both datasets, \textsc{FedPG} requires an exchange of substantially smaller model sizes compared to other baselines. For the UNSW-NB15 datasets, the model size of \textsc{FedPG} is approximately 6.7 and 10 times more efficient than FedAE and FedBiGAN, respectively, while achieving better performance. Meanwhile, FedAE-2 and FedBiGAN-2 might perform moderately better, but require a 10 and 15 times larger models than \textsc{FedPG}. 
A similar efficiency is observed for the TON-IoT dataset, where \textsc{FedPG} outperforms by factors of up to 20 times compared to other methods. These results hold considerable significance in the context of FL, where the reduction of bandwidth utilization is a paramount concern. This evidence underscores the ability of \textsc{FedPG} to strike an effective balance between reliable anomaly detection and optimal resource usage, thus bolstering its suitability for real-world NIDS applications.

\subsubsection{ Memory and Time Complexity}
We analyze all models through the lens of GPU memory consumption (measured in a percentage) and wall-clock time for inference (measured in a microsecond - $\mu s$), a perceptible trade-off emerges between these two metrics. As depicted in Fig. \ref{fig:memory}, \textsc{FedPG} and \textsc{FedPE} are relatively memory-efficient, consuming around 9.03\% and 8.01\% of GPU memory during training, respectively. In contrast, the more sophisticated models, FedAE, FedAE-2, FedBiGAN, and FedBiGAN-2, are more resource-demanding, utilizing larger amounts of memory. Despite the smaller memory footprint, \textsc{FedPG} and \textsc{FedPE} do not compromise on the inference speed, clocking an impressive inference wall-clock time of approximately 7$\mu s$. This slightly outperforms the inference time taken by other non-linear methods. Thus, the superior balance struck by \textsc{FedPG} and \textsc{FedPE} between GPU memory usage and wall-clock time solidifies their efficiency, making them well-suited for resource-restricted environments.

\section{Conclusion}
\label{sec:conclusion}
In this study, we analyze the convergence of the federated PCA framework on the Grassmann manifold, thereby enhancing efficient unsupervised anomaly detection in IoT networks with realistic datasets. The proposed framework empowers IoT devices to collaboratively create a profile of normal network traffic while abstaining from the exchange of local data. Specifically, we devise \textsc{FedPE}, an innovative federated algorithm that approximates PCA utilizing the principle of ADMM and is accompanied by theoretical guarantees. We further conceptualize the federated PCA via Grassmannian optimization and propose another algorithm, \textsc{FedPG}. This algorithm is geared towards ensuring a swift training process and early anomaly detection, even within computing environments that are resource-constrained. Through our experimental evaluations, we establish the adaptability of \textsc{FedPE} and \textsc{FedPG} to the IoT environment and demonstrate their superior performance over conventional approaches when tested on the UNSW-NB15 and TON-IoT datasets.

% \section{Acknowledgement}
% This work was partly supported by the Institute of Information \& communications Technology Planning \& Evaluation (IITP) grant (No.2019-0-01287 \& No.RS-2022-00155911), by the National Research Foundation of Korea(NRF) (No. RS-2023-00207816), by IITP grant funded by MSIT (No.2021-0-02068, Artificial Intelligence Innovation Hub), by USyd SOAR Prize 2022-2023, and by USyd-UofG Ignition Grant 2023-2024. The work of Long Tan Le was supported by the PhD funding from the Vingroup Science and Technology Scholarship Program for Overseas Study.

%Long Tan Le acknowledges the PhD funding support from the Vingroup Science and Technology Scholarship Program, which contributed to this work.

%\clearpage
%\newpage
\bibliographystyle{IEEEtran}
\bibliography{references} % Path to your References.bib file
\section{Biography Section}

%\vspace{11pt}
\begin{IEEEbiography}[{\includegraphics[width=1in,height=1.25in,clip,keepaspectratio]{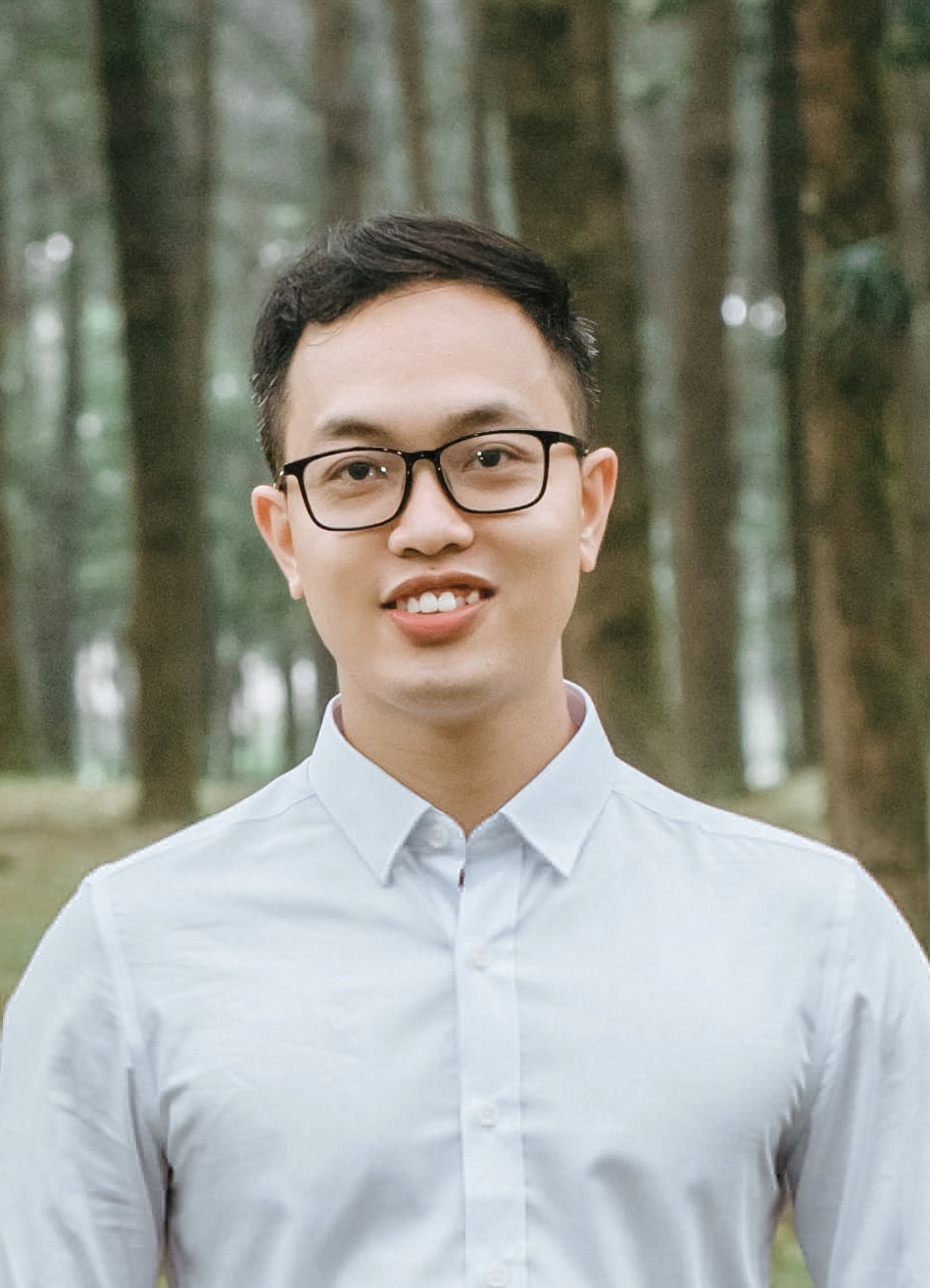}}]{Tung-Anh Nguyen} received his B.E. degree in Electronics and Telecommunications from Hanoi University of Science and Technology (HUST), Vietnam, in 2016. He received his MSC degree in Computer and Information Science from Korea University, Korea, in 2019. Currently, he is pursuing a PhD degree in Computer Science at DUAL group, Faculty of Engineering, the University of Sydney (USyd), NSW, Australia. His research interests include Distributed Machine Learning, Federated Learning, Blockchain and Optimization for Distributed Systems.
\end{IEEEbiography}
%\vspace{11pt}
\begin{IEEEbiography} [{\includegraphics[width=1in,height=1.25in,clip,keepaspectratio]{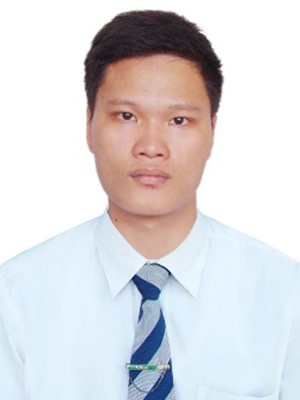}}]{Long Tan Le} received B.E. degree with honours in Computer Engineering from Ho Chi Minh City University of Technology (HCMUT) in 2018. He was a researcher at the Faculty of Computer Science and Engineering, HCMUT from 2018 to 2021. Currently, he is pursuing a Ph.D degree in Computer Science at the Faculty of Engineering, the University of Sydney (USyd). His research interests include Machine Learning, Federated Learning, and their applications for next-generation networking.
\end{IEEEbiography}
%\vspace{11pt}
\begin{IEEEbiography}[{\includegraphics[width=1in,height=1.25in,clip,keepaspectratio]{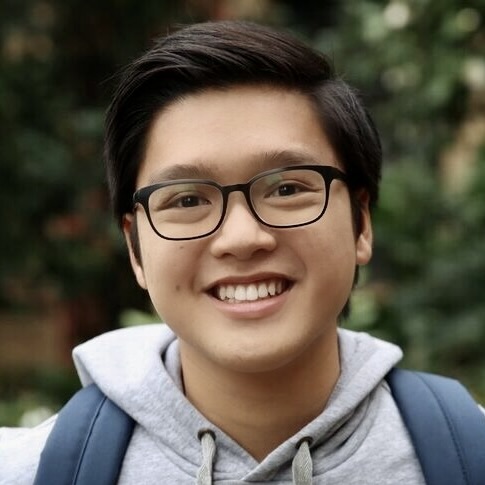}}]{Tuan Dung Nguyen} received his B.S. from the University of Melbourne and M.Phil. from the Australian National University, both in computer science. He is currently pursuing a Ph.D. in computer and information science at the Uniersity of Pennsylvania, where he is a part of the Computational Social Science Lab. His research interests span machine learning, convex optimization, distributed computing and social network analysis.

\end{IEEEbiography}
%\vspace{11pt}
\begin{IEEEbiography}[{\includegraphics[width=1in,height=1.25in,clip,keepaspectratio]{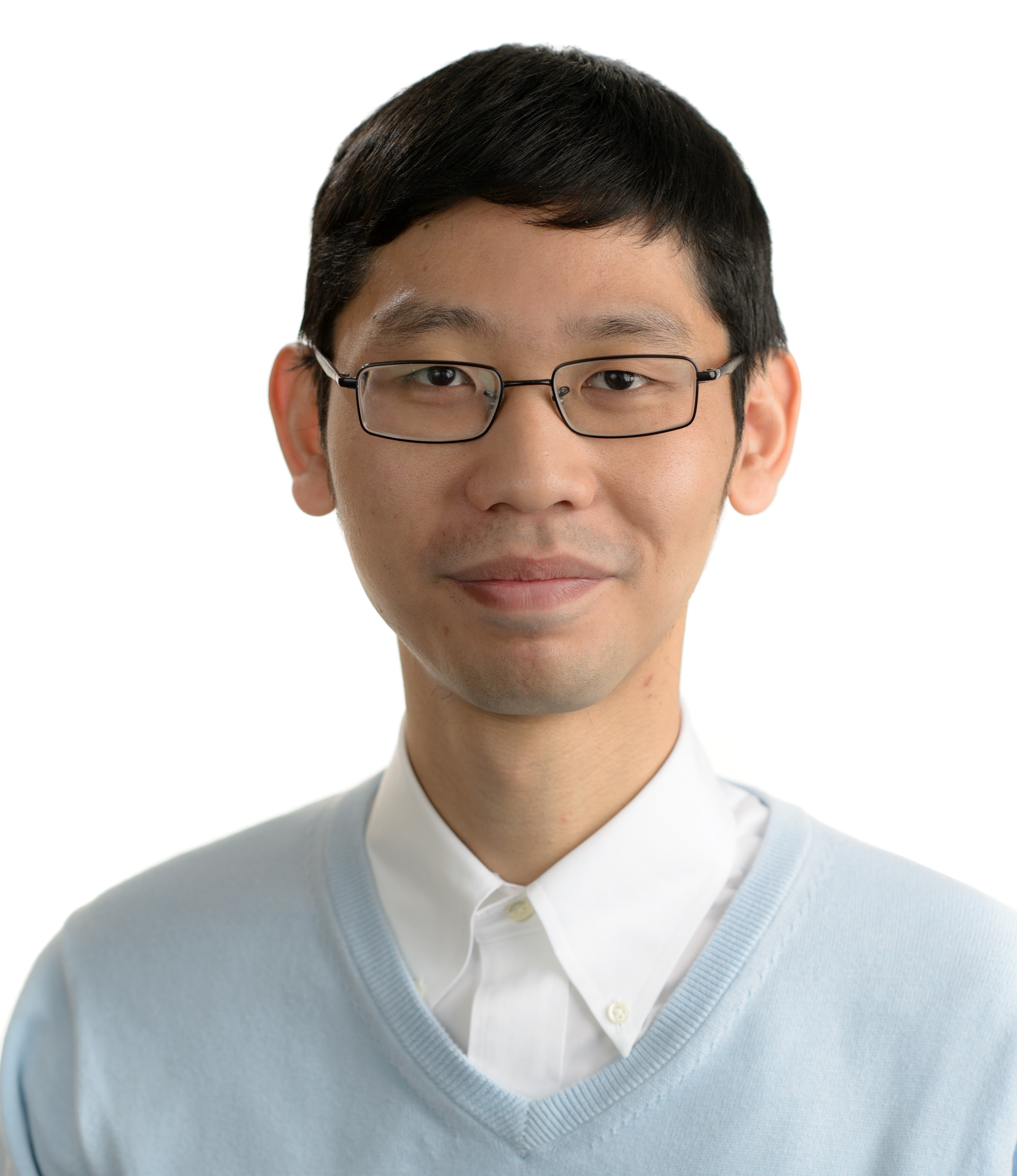}}]{Wei Bao} (Member, IEEE) received the BE degree in communications engineering from the Beijing University of Posts and Telecommunications, Beijing, China, in 2009, the MASc degree in electrical and computer engineering from the University of British Columbia, Vancouver, Canada, in 2011, and the PhD degree in electrical and computer engineering from the University of Toronto, Toronto, Canada, in 2016. He is currently a senior lecturer with the School of Computer Science, University of Sydney, Sydney, Australia. His research covers the area of network science, with particular emphasis on Internet of things, mobile computing, edge computing, and distributed learning. He received the Best Paper Awards in ACM International Conference on Modeling, Analysis and Simulation of Wireless and Mobile Systems (MSWiM) in 2013 and 2019 and IEEE International Symposium on Network Computing and Applications (NCA) in 2016.
\end{IEEEbiography}
%\vspace{11pt}
\begin{IEEEbiography}[{\includegraphics[width=1in,height=1.25in,clip,keepaspectratio]{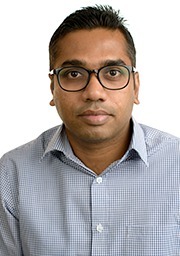}}]{Suranga Seneviratne} (Senior Member, IEEE) received the bachelor’s degree from University of Moratuwa, Sri Lanka, in 2005, and the Ph.D. degree from University of New South Wales, Australia, in 2015. He is a Lecturer of Security with the School of Computer Science, The University of Sydney. His current research interests include privacy and security in mobile systems, AI applications in security, and behavior biometrics. Before moving into research, he worked nearly six years in the telecommunications industry in core network planning and operations
\end{IEEEbiography}
%\vspace{11pt}
\begin{IEEEbiography}[{\includegraphics[width=1in,height=1.25in,clip,keepaspectratio]{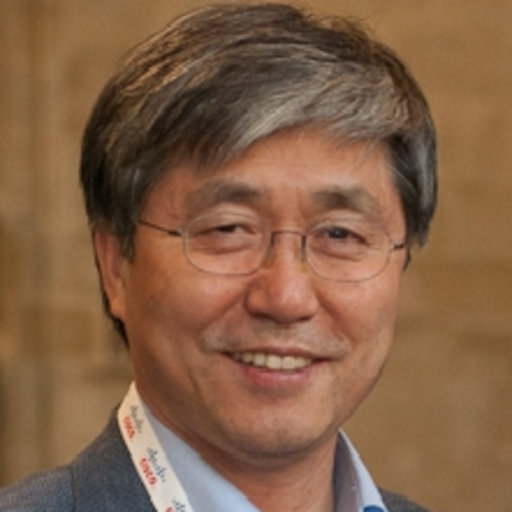}}]{Choong Seon Hong} (S’95-M’97-SM’11) received the B.S. and M.S. degrees in electronic engineering from Kyung Hee University, Seoul, South Korea, in 1983 and 1985, respectively, and the Ph.D. degree from Keio University, Tokyo, Japan, in 1997. In 1988, he joined KT, Gyeonggi-do, South Korea, where he was involved in broadband networks as a member of the Technical Staff. Since 1993, he has been with Keio University. He was with the Telecommunications Network Laboratory, KT, as a Senior Member of Technical Staff and as the Director of the Networking Research Team until 1999. Since 1999, he has been a Professor with the Department of Computer Science and Engineering, Kyung Hee University. His research interests include future Internet, intelligent edge computing, network management, and network security. Dr. Hong is a member of ACM, IEICE, IPSJ, KIISE, KICS, KIPS, and OSIA. He was an Associate Editor of the IEEE TNSM and the IEEE JSAC.
\end{IEEEbiography}
%\vspace{11pt}
\begin{IEEEbiography}[{\includegraphics[width=1in,height=1.25in,clip,keepaspectratio]{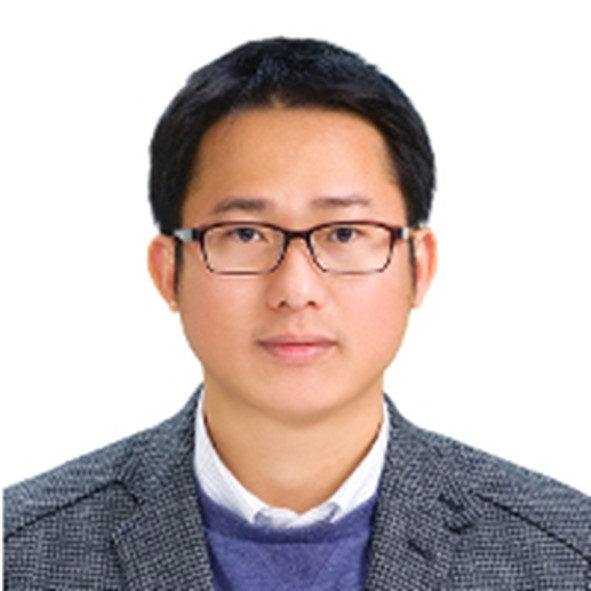}}]{Nguyen H. Tran} (S’10-M’11-SM’18) received BS and PhD degrees (with best PhD thesis award), from HCMC University of Technology and Kyung Hee University, in electrical and computer engineering, in 2005 and 2011, respectively. Dr Tran is an Associate Professor at the School of Computer Science, The University of Sydney. He was an Assistant Professor with Department of Computer Science and Engineering, Kyung Hee University, from 2012 to 2017. His research group has special interests in Distributed compUting, optimizAtion, and machine Learning (DUAL group). He received several best paper awards, including IEEE ICC 2016 and ACM MSWiM 2019. He receives the Korea NRF Funding for Basic Science and Research 2016-2023, ARC Discovery Project 2020-2023, and SOAR award 2022-2023. He has been on the Editorial board of several journals such as IEEE Transactions on Green Communications and Networking (2016-2020), IEEE Journal of Selected Areas in Communications 2020, and IEEE Transactions on Machine Learning in Communications Networking (2022-).
\end{IEEEbiography}
%\vspace{11pt}
\vfill
%\section*{Acknowledgments}
%This should be a simple paragraph before the References to thank those %individuals and institutions who have supported your work on this article.

%{\appendices
%\section*{Proof of the First Zonklar Equation}
%Appendix one text goes here.
% You can choose not to have a title for an appendix if you want by leaving the argument blank
%\section*{Proof of the Second Zonklar Equation}
%Appendix two text goes here.}

\end{document}